\documentclass[preprint,review,10pt]{elsarticle}
\usepackage{amssymb,geometry,xcolor,amsmath,bbm,multirow,bbold,mathtools,amssymb,graphics,graphicx,epstopdf,ntheorem,diagbox,enumitem,diagbox}
\usepackage{xr}
\usepackage{booktabs} 
\newtheorem{lemma}{Lemma}
\newtheorem{theorem}{Theorem}

\newtheorem{proposition}{Proposition}

\usepackage[ruled,linesnumbered]{algorithm2e}
\newcommand{\rF}{\rm F}
\newcommand{\argmin}{\arg\min}
\newcommand{\argmax}{\arg\max}

\journal{Pattern Recognition}

\begin{document}

\begin{frontmatter}

\title{Regularized Projection Matrix Approximation with Applications to Community Detection}

\author[bnu]{Zheng Zhai}
\author[bnu]{Jialu Xu}
\author[bnu]{Mingxin Wu}

\affiliation[bnu]{organization={Beijing Normal University},
            addressline={No.18 Jinfeng Road}, 
            city={Zhuhai},
            postcode={519087}, 
            state={Guangdong},
            country={China}}
\author[HH]{Xiaohui Li\corref{hh}}

\affiliation[HH]{organization={Yantai University},
addressline={No.30 Laishan Qingquan Road}, 
city={Yantai},
postcode={264005}, 
state={Shandong},
country={China}}
            
\cortext[hh]{Corresponding author: Xiaohui Li (lxhmath@zju.edu.cn).}

\begin{abstract}
This paper introduces a regularized projection matrix approximation framework designed to recover cluster information from the affinity matrix. The model is formulated as a projection approximation problem, incorporating an entry-wise penalty function. We investigate three distinct penalty functions, each specifically tailored to address bounded, positive, and sparse scenarios. To solve this problem, we propose direct optimization on the Stiefel manifold, utilizing the Cayley transformation along with the Alternating Direction Method of Multipliers (ADMM) algorithm. Additionally, we provide a theoretical analysis that establishes the convergence properties of ADMM, demonstrating that the convergence point satisfies the KKT conditions of the original problem. Numerical experiments conducted on both synthetic and real-world datasets reveal that our regularized projection matrix approximation approach significantly outperforms state-of-the-art methods in clustering performance.

\end{abstract}

%

%


\end{frontmatter}

\section{Introduction}
Community detection is a crucial problem in unsupervised learning that has garnered attention from researchers across various disciplines, including mathematics, statistics,  physics, and social sciences. The objective is to partition $n$ data points into $K$ groups based on their pairwise similarities, represented by a similarity matrix $A\in {\mathbb R}^{n×n}$. A prevalent approach to tackle this problem involves first deriving a lower-dimensional representation of the data from $A$. Subsequently, a clustering algorithm such as $k$-means or the EM algorithm is applied to identify the clusters. 
The success of this method depends on the quality of the data representation and the accuracy of the computational methods used for $A$.


A popular approach for cluster identification is to utilize the top $K$ eigenvectors of matrix $A$, as employed in spectral clustering~\cite{white2005spectral,von2007tutorial}. Identifying these eigenvectors is equivalent, up to rotations, to determining the subspace spanned by them. This, in turn, is analogous to solving the following projection matrix approximation problem:
\begin{align}
\widehat X=\argmin_{X\in\mathcal{P}_{K}}\|A-X\|^2_{\rF},\label{equ:1}
\end{align}
where $\mathcal{P}_{K}\subseteq\mathbb{R}^{n\times n}$ is the set of rank-$K$ projection matrices. Thus, the effectiveness of spectral clustering is highly dependent on the quality of the projection matrix approximation. 

Similar to the affinity matrix, the projection matrix also encapsulates class information. Consider an extreme scenario where clustering information is known beforehand, with $K$ classes, and each class consists of $\{n_k, k=1,...,K\}$ samples. In this setup, the affinity matrix $A$ is defined such that $A_{i,j}=1$ if and only if samples $i$ and $j$ belong to the same group, and $A_{i,j}=0$ otherwise. Solving the corresponding eigenvalue problem provides the solution for \eqref{equ:1} as follows:
\begin{equation}\label{assignP}
\widehat{X}_{i,j} = \left\{
\begin{array}{cc}
  0,  & A_{i,j}=0,\\
  1/n_k,    & A_{i,j}=1,\ c(i)=c(j)=k .
\end{array}
\right.
\end{equation}
where $c(\cdot)$ is the function which maps the sample index to its corresponding cluster number. It can be verified that $\widehat{X}$ is a low rank, non-negative, sparse projection matrix satisfying $\widehat{X}^2 =\widehat{X}$. If we shuffle the indices and rearrange them according to the cluster indicator, we obtain a new projection matrix, which can be written as
\begin{equation}\notag
P\widehat{X} P^T =\left[ \begin{array}{cccc} \frac{1}{n_1} {\bf 1}_{n_1\times n_1} & \bf 0& ...&\bf 0\\
\bf 0 & \frac{1}{n_2} {\bf 1}_{n_2\times n_2}&  ... & \bf 0\\
...&...&...&...\\
\bf 0 & ...&... &\frac{\bf 1}{n_K} {\bf 1}_{n_K\times n_K}
\end{array} \right],
\end{equation}
where ${\bf 1}_{n_k\times n_k}\in \mathbb R^{n_k\times n_k}$ is a squared matrix with all elements equal to $1$, and $P$ is a permutation matrix. We can observe that both $P\widehat{X} P^T$ and $\widehat{X}$ contain exactly of $\sum_k n_k^2$ nonzero elements.

Inspired by the optimal solution presented in equation \eqref{assignP}, we propose that an effective projection matrix for clustering should exhibit certain distinctive properties such as non-negativity, boundedness, and a degree of sparsity. In this work, we introduce a uniform regularized projection problem approximation framework aimed at boosting clustering performance. Specifically, we study the entry-wise regularized projection approximation problem to restrict the entries of the projection matrix within a reasonable scope by considering:
\begin{align}\label{RP} 
\min_{X\in\mathcal{P}_K\cap {\cal C}}\|A-X\|_{\rF}^2.
\end{align}
where $\mathcal{P}_K$ is the set consisting of all rank-$K$ projection matrices and $\cal C$ is some constraint set such as the pairwise bounded restriction $\{X|X_{i,j}\in [\alpha,\beta],\forall i,j\}$, the non-negative constraint $\{X|X_{i,j}\geq 0,\forall i,j\}$ or even the sparsity constraint such as $\{X|\|X\|_0\leq\eta\}$. Due to the simultaneously 
existence of the projection and extra constraint $\cal C$, it is challenging to directly solve \eqref{RP}. To address this difficulty, this paper proposes a regularized projection matrix approximation (RPMA) problem in \eqref{Convex} as an approximation to the above strict constrained problem, which is defined by
\begin{align}\label{Convex} 
   \min_{X\in\mathcal{P}_K}\|A-X\|_{\rF}^2+\lambda\sum_{ij}g(X_{ij}),\ \ g\in {\cal G}.
\end{align}
where $\lambda$ serves as a non-negative tunable parameter, governing the trade-off between approximation error and entry-wise constraint. We restrict $\cal G$ as the class of positive, convex functions whose derivatives are continuous and satisfy the Lipschitz condition, i.e.,
\[
\begin{aligned}
{\cal G} := \{g| g\in {C}^{(1)},\ g(\cdot)\geq 0,\ g''(\cdot) \geq 0, 
|g'(x_1)-g'(x_2)| \leq \ell |x_1-x_2|, \forall x_1,x_2\}.
\end{aligned}
\]
where ${C}^{(1)}$ represents the class of functions whose derivatives are continuous and $\ell$ is a positive constant. The construction of $\cal G$ is driven by two key considerations: Firstly, it must be expansive enough to adequately encapsulate our objectives as a penalty term. Secondly, it should adhere to specific properties that facilitate algorithm development and validate algorithm's convergence behaviors.

We solve the optimization problem in \eqref{Convex} from two distinct and complementary perspectives: the first focuses on optimizing over the Stiefel manifold, while the second involves separating the constraints via introducing an axillary variable and applying the Alternating Direction Method of Multipliers (ADMM). 

{\it Optimizing on Stiefel manifold:}
Noting that any rank-$K$ projection matrix $X \in \mathbb{R}^{n \times n}$ can be expressed as $X = UU^T$, where $U\in \mathbb{R}^{n \times K}$ is an orthonormal matrix satisfying $U^T U = I_K$, we reformulate \eqref{Convex} into an equivalent form:
\[
\min_{U^T U = I_k} \|A - UU^T\|_{\rm F}^2 + \lambda \sum_{ij} g(\{UU^T\}_{i,j}).
\]
To solve it, we parameterize the univariable curve $U(t)$ on the Stiefel manifold using the Cayley transformation and employ curvilinear search method to find the optimal solution along this curve.

{\it Constraints Separation via {\rm ADMM}:}
We note that the primary challenge in solving \eqref{Convex} stems from the simultaneous imposition of the rank-$K$ projection constraints $X \in {\cal P}_K$ and the incorporation of the penalty term $g(X_{i,j})$.
To manage these issues, we propose separating the constraints by introducing an auxiliary variable $Y$, replacing the penalty term with respect to $g(Y_{i,j})$. We then employ a dual variable strategy to restrict the distance between $X$ and $Y$ using the Alternating Direction Method of Multipliers (ADMM). This approach allows us to handle the projection and penalty terms more efficiently by alternating between updating $X$ and $Y$, while controlling their proximity via the dual variables.

Our contributions are threefold. First, we introduce a novel regularized projection matrix approximation framework aimed at improving the classical spectral clustering method. This framework incorporates three distinct types of penalty functions, each tailored for specific scenarios: bounded, nonnegative, and sparse. Second, we propose two algorithmic approaches: one that directly optimizes on the Stiefel manifold using the Cayley transformation, and another based on the ADMM algorithm. We also rigorously analyze the KKT conditions and demonstrate that the convergence point of the ADMM algorithm satisfies the KKT conditions of the original problem. Third, we validate the effectiveness of our approach through extensive experiments on both synthetic and real-world datasets, showcasing its practical utility and robustness.
\subsection{Related works}

Low-rank matrix optimization with additional structural constraints is a prevalent problem in machine learning and signal processing~\cite{zhang2012low,zhang2012inducible}. The objective is to find the best low-rank matrix approximation that also satisfies certain structural constraints, such as non-negativity, symmetry, boundedness~\cite{thanh2022bounded}, simplex~\cite{abdolali2021simplex}, and sparsity~\cite{pan2019generalized}. There are two primary approaches to achieve this target.
Firstly, these constraints can be enforced via matrix factorization with explicit constraints. Examples include non-negative matrix factorization~\cite{lee1999learning}, semi-nonnegative matrix factorization~\cite{ding2008convex}, bounded low-rank matrix approximation~\cite{kannan2012bounded}, and bounded projection matrix approximation~\cite{BPMA}.
Secondly, constraints can also be achieved via the soft-regularization term approach. This method includes techniques like simultaneously low-rank and sparse matrix approximation~\cite{zhang2022graph,richard2012estimation}.
However, these works such as~\cite{nie2016constrained,ji2011robust} primarily focus on seeking a low-rank matrix as a low-dimensional embedding of the input similarity matrix. In contrast, our work diverges from this goal. Instead of learning an embedding, we aim to recover a high-quality group connection matrix through regularized projection matrix approximation.

\section{Constraints and Penalties }

In this section, we study different forms of penalty function $g(\cdot)$ for achieving the various targeted constrains such as non-negativity, boundedness, and sparsity. 


\subsection{Bounded Penalty}
Based on \eqref{assignP}, it's evident that the low-rank approximation $X$ of an assignment matrix must be confined within $[0,1/\min_k n_k]$. Hence, it's imperative for us to learn a bounded rank-$K$ projection matrix from $A$. In a general form, we address the problem as finding a projection matrix where each element falls within $[\alpha,\beta]$. One straightforward approach is to introduce the indicator function $I(x)$, defined such that $I(x)=0$ when $x$ lies in $[\alpha,\beta]$, and $I(x)=+\infty$ otherwise. However, this function's lack of continuity poses difficulties when employing differentiable tools for analysis.
To mitigate this, we introduce a smooth, convex function $g_{\alpha,\beta}(z)$ as a relaxation of the bounded indicator function, which is defined as 
\begin{equation}\label{bb}
g_{\alpha,\beta}(z) = (\min\{z-\alpha,0\})^2+(\min\{\beta-z,0\})^2.
\end{equation}

The construction of $g_{\alpha,\beta}(z)$ reveals that it solely penalizes $z$ when it lies outside the range $[\alpha,\beta]$.
It can be observed that as $\lambda$ approaches $+\infty$, $\lambda g_{\alpha,\beta}(z)$ exhibits behavior akin to that of the indicator function. Moreover, this function is convex, and its derivative is Lipschitz continuous, satisfying $|g'_{\alpha,\beta}(x_1)-g'_{\alpha,\beta}(x_2)|\leq 2|x_1-x_2|$.


\subsection{Non-negativity Penalty}

The non-negativity constraint can be viewed as a one-side boundedness requirement by setting $\alpha=0,\beta=+\infty$. Thus, we can consider a non-negativity penalty function as a special case of the bounded penalty in \eqref{bb}. Therefore,
we introduce this non-negative penalty function to regularize the projection matrix, which is defined as
\begin{equation}\label{nn}
 g(z) :=  (\min\{z,0\})^2.
\end{equation}

Similarly to the bounded penalty, the function $g(z)$ only exerts an effect when $z<0$. Furthermore, as $\lambda$ tends to infinity, $\lambda g(z)$ exhibits behavior akin to the indicator function $I(z)$, where $I(z)=0$ if and only if $z \geq 0$, and $I(z)=+\infty$ otherwise. Additionally, $g(z)$ is nonnegative and convex, and its derivative is Lipschitz continuous, satisfying $|g'(x_1)-g'(x_2)| = 2|x_1-x_2|$. 

\vspace{-2mm}
\subsection{Sparsity Penalty}\label{sparse case}

\begin{figure}[t]
\centering
\includegraphics[width=\linewidth]{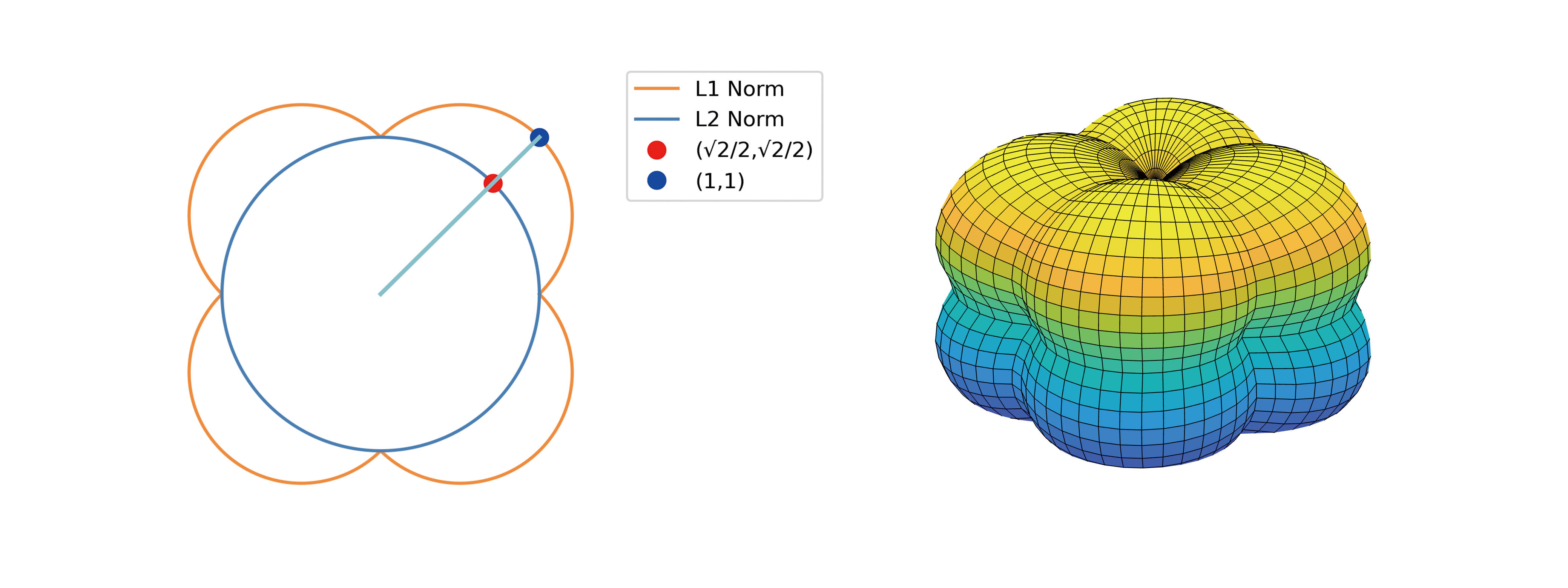}
\vspace{-10mm}
\caption{Illustration demonstrating why minimizing the $\ell_1$ norm leads to a sparse solution. The left diagram shows the curve
  $r(x_1,x_2)$ with $r = |x_1|+|x_2|$ and $x_1^2+x_2^2=1$, while the right diagram depicts the surface $r(x_1,x_2,x_3)$ with $r=|x_1|+|x_2|+|x_3|$ and $x_1^2+x_2^2+x_3^2=1$.\label{l1l2}
}
\end{figure}

The Lasso penalty~\cite{tibshirani1996regression} encourages sparsity by inducing certain coefficients or parameters in a model to become exactly zero. The rational for minimizing the $\ell_1$ norm yields a sparse solution can be illustrated by an example. In Figure \ref{l1l2}, we provide a unit circle to represent the data $x\in \mathbb R^2$ (left diagram) and $x\in \mathbb R^3$ (right diagram) with fixed $\ell_2$ norm, and also plot the data by re-scaling with the $\ell_1$ as $y=\|x\|_1x$. From Figure \ref{l1l2}, we observe that $\min_{\|x\|_2=1}\|x\|_1$ achieves four optimal solutions: $(1,0),(-1,0),(0,1),(0,-1)$ when $x\in {\mathbb R}^2$ in the left diagram, and six optimal solutions: $(1,0,0),(-1,0,0),(0,1,0),(0,-1,0),(0,0,1)$ and $(0,0,-1)$ when $x\in {\mathbb R}^3$ in the right diagram. Since any rank-$K$ projection matrix has a fixed Frobenius norm equal to $\sqrt{K}$, we speculate that $\min_{X\in {\cal P}_{K}} \|X\|_1$ will also yield a sparse solution, similar to the examples in $\mathbb R^2$ and $\mathbb R^3$.

Motivated by the example in Figure \ref{l1l2}, it is natural to set $g(z)$ as the absolute value function $|z|$ to regularize the projection approximation problem. Consequently, the summation of $|X_{i,j}|$ leads to the Lasso penalty function $\|X\|_1$ such that $\sum_{i,j} g(X_{i,j}) = \|X\|_1$. However, the derivative of the absolute value function is non-differentiable at $0$ and does not satisfy the Lipschitz continuous property. This can be verified as follows: for any $x_1\rightarrow 0^-, x_2\rightarrow 0^+$, we have $|g'(x_1)-g'(x_2)|=2$, while $|x_1-x_2|$ can be arbitrarily small. Therefore, the absolute function does not belong to $\cal G$. 

To address this, we can approximate the absolute function by introducing
the Huber loss function~\cite{sun2020adaptive,huang2021robust}, which smooths the absolute value function with a quadratic function in the neighborhood of $0$. The Huber loss function is defined by 
\begin{equation}\label{hl}
g_\delta(x) = \left\{
\begin{array}{ll}
      \frac{x^2}{2\delta}, & |x|\leq \delta, \\
      |x|-\frac{\delta}{2}, & {\rm otherwise},
\end{array}
\right.
\end{equation}
where $\delta$ is a positive threshold parameter. It can be verified that Huber function retains the advantages of the absolute value function while ensuring both differentiability and Lipschitz continuity at $x=0$.

It is worth mentioning that as $\delta \rightarrow 0^+, g_\delta(x)$ uniformly converges to $|x|$. Thus, the Huber loss function can induce sparse solutions similarly to the Lasso penalty when $\delta$ is small enough. We propose using the sum of the Huber loss function, $\sum_{i,j}g_\delta (X_{i,j})$, as a substitute for the Lasso penalty $\|X\|_1$ to learn a sparse projection approximation by adopting a sufficiently small $\delta$. Compared to the absolute value function, the Huber loss function possesses certain advantages, such as smoothness and Lipschitz continuity, demonstrated by $|g'_\delta(x_1) - g'_\delta(x_2)| \leq \frac{1}{\delta} |x_1 - x_2|$.

\section{First-order optimal condition}

For notational convenience, let the projection matrix be denoted as $X := UU^T$. 
Therefore, the problem in \eqref{Convex} can be equivalently reformulated as:
\begin{equation}\label{equ}
\min_{U^TU = I_d} F(U) = \|A - UU^T\|_{\rm F}^2 + \lambda \sum_{i,j} g(X_{i,j}).
\end{equation}

This is a nonlinear optimization problem constrained by the Stiefel manifold. The optimal solution to the problem 
$\min_{U^TU=I_d} F(U)$ is not unique. Specifically, if $U$  is an optimal solution for $F(U)$, then for any orthogonal matrix $Q\in {\mathbb R}^{K\times K}$, we have $F(UQ) = F(U).$ Therefore, we approach the optimization of $U$ from a subspace perspective: if $U'$ and  $U$ span the same subspace, we consider them equivalent, denoted by $U'\sim U$. Next, we explore the first order condition for \eqref{equ}.
\subsection{KKT Condition} \label{KKTC}
The Euclidean gradient of $F(U)$ with respect to $U$ is given by $\nabla_U F(U) = -4AU + \lambda M$, where $M$ represents the gradient of $\sum_{i,j} g(X_{i,j})$ with respect to $U$. Specifically, we compute the $i,j$-th entry for $M_{i,j}$:
\begin{equation}\label{deriv}
M_{i,s} = 
\frac{\partial \sum_{ij}g(X_{i,j})}{\partial U_{i,s}} = \frac{\partial \sum_{i\neq j} g(X_{i,j})}{\partial U_{i,s}}+\frac{\partial \sum_{i} g(X_{i,i})}{\partial U_{i,s}} = 2(\sum_{j\neq i} g'(X_{i,j})U_{j,s} + g'(X_{i,i})U_{i,s}).
\end{equation}
In matrix form, we can express $M$ as: $M = 2G U$, where $G$ represents a symmetric matrix of the form
\begin{equation}\label{G}
\quad G :=
\left[
\begin{array}{cccc}
 g'(X_{1,1}) & g'(X_{1,2}) & \cdots & g'(X_{1,n}) \\
g'(X_{2,1}) &  g'(X_{2,2}) & \cdots & g'(X_{2,n}) \\
\vdots & \vdots & \ddots & \vdots \\
g'(X_{n,1}) & g'(X_{n,2}) & \cdots &  g'(X_{n,n})
\end{array}
\right].
\end{equation}
After deriving the gradient with respect to $U$, we can obtain the first-order optimality condition. First, we define the the Lagrangian for equation \eqref{equ} as follows:
\begin{equation}\label{Lag}
L(U, \Lambda) = \|A - UU^T\|_{\rm F}^2 + \lambda \sum_{i,j} g(X_{i,j}) + \langle U^TU-I_d,\Lambda\rangle.
\end{equation}
where $\Lambda$ can be selected as any symmetric matrix, owing to the symmetry of $U^TU-I_d$. Additionally, $\Lambda$ possesses an eigenvalue decomposition given by $\Lambda = Q\Lambda'Q^T$, where $Q$ is an orthogonal matrix and $\Lambda'$ is a diagonal matrix. If we define a new orthonormal matrix $\hat{U} = UQ$, the Lagrangian in equation \eqref{Lag} can be expressed as follows:
\[
L(\hat{U}, \Lambda_d) = \|A - \hat{U}\hat{U}^T\|_{\rm F}^2 + \lambda \sum_{i,j} g(X_{i,j}) + \langle \hat{U}^T\hat{U}-I_d,\Lambda_d\rangle.
\]
Thus, we can conclude that $\Lambda$ in \eqref{Lag} can be chosen as a diagonal matrix, and moving forward, we will treat $\Lambda$ as such. By taking the derivatives of $\Lambda$ in $L(U,\Lambda_d)$ with respect to both $U$ and $\Lambda_d$ and set them to $\bf 0$, we derive the KKT condition for \eqref{equ} as follows:
\begin{equation}\label{nonlinear_eign}
\left\{
\begin{aligned}
&(2 A-\lambda G)U  = U\Lambda_d ,\\
&U^TU=I_d.
\end{aligned} 
\right.
\end{equation}

Thus, the columns of $U$ corresponds to the eigenvector of $2 A-\lambda G$ with the diagonal elements in $\Lambda_d$ stands for the eigenvalues.
Because  $G$  is a matrix that depends on $U$ via $X$, and $\Lambda_d$ is a diagonal matrix, we refer to \eqref{nonlinear_eign} as a nonlinear eigenvalue problem. The nonlinearity arises from the fact that  $G$  evolves based on $U$, creating a feedback loop between $G$ and $U$, which complicates the problem compared to traditional linear eigenvalue problems.  
\subsection{Geometric Interpretation}
In what follows, we give a deeper exploration of the KKT condition and also provide a geometric interpretation of the first-order optimal condition from the perspective of tangent space. 

First, we eliminate the variable $\Lambda_d$. By multiplying $U^T$ on both sides of \eqref{nonlinear_eign}, we can solve $\Lambda_d$. Specifically, this yields: $\Lambda_d=\Lambda_d^T = -\frac{1}{2}U^T \nabla_U F(U)$. Substituting  $\Lambda_d = -\frac{1}{2}\nabla_U^T F(U)U$ back into the first condition of \eqref{nonlinear_eign}, the KKT condition can be rewritten as:
\begin{equation}\label{equ_nonlinear_eign}
\left\{
\begin{aligned}
&\nabla_U F(U)  - U \nabla_U^T F(U) U = \bf 0,\\
&U^TU=I_d.
\end{aligned} 
\right.
\end{equation}
Next, we give the geometric interpretation of \eqref{equ_nonlinear_eign}. Notice that:
\begin{equation}\label{tangent_expen}
\nabla_U F(U) = U U^T \nabla_U F(U)  + U^\perp ({U^\perp})^T \nabla_U F(U).
\end{equation}
By substituting \eqref{tangent_expen} into \eqref{equ_nonlinear_eign}, we have:
\begin{equation}\label{U0}
\underbrace{U  (U^T \nabla_U F(U) -  \nabla_U^T F(U) U) + U^\perp ({U^\perp})^T  \nabla_U F(U) }_{\nabla_c F(U)}= \bf 0.
\end{equation}

Since $U$ and $U^\perp$ are orthogonal, \eqref{U0} implies that both of the coordinate matrices satisfy $U^T \nabla_U F(U) -  \nabla_U^T F(U) U=\bf 0$ and $({U^\perp})^T  \nabla_U F(U)=\bf 0$. 

\begin{proposition} \label{consistency}
    Denote the the canonical metric $\langle \cdot,\cdot \rangle_c$ by $\langle A,B \rangle_c := {\rm trace} (A^T (I-\frac{1}{2}UU^T)B), \forall  A,B\in T_U.$ Let  $P_{T_U}(\nabla F(U))$ denote the projection of $\nabla F(U)$ onto $T_U$. Then, for any $V\in T_U$, we have:
    \[
    \langle\nabla_c F(U),V\rangle_c = \langle \nabla F(U), V \rangle 
 = \langle P_{\rm T_U}(\nabla F(U)), V \rangle,
    \]
    where the projection operator is defined as $P_{\rm T_U}(W) = U  \frac{(U^T W -  W^T U)}{2} + U^\perp ({U^\perp})^T W$. 
\end{proposition}

Proposition \ref{consistency} captures the relationship between the canonical gradient and its projection within the tangent space, ensuring consistency across different metrics. It is also important to note that $P_{T_U}(\nabla F(U))$ can be derived from the following optimization problem :
\begin{equation}\label{op_proj}
(\widehat{Y},\widehat{Z}) = \min_{Y^T=-Y,Z} \|UY+U^\perp Z- \nabla_U F(U)\|_{\rm F}^2,
\end{equation}
where $UY+U^\perp Z$ (with $Y$ being skew-symmetric) represents the general form of vectors in $T_U$.
It is evident that the optimization problem \eqref{op_proj} yields the closed-form solution given by $\widehat{Y} = \frac{1}{2}(U^T \nabla_U F(U) -  \nabla_U^T F(U) U)$ and $\widehat{Z} = ({U^\perp})^T  \nabla_U F(U)$.

Therefore, the KKT condition $\nabla_c F(U) = \bf 0$ implies that $P_{T_U}(\nabla F(U)) = \bf 0$. This means that the Euclidean gradient, when projected onto the tangent space $T_U$, vanishes, indicating that there are no directions of feasible descent within that tangent space.

\section{Algorithm}
We propose two distinct algorithms to solve our model. The first approach involves direct optimization on the Stiefel manifold using the Cayley transformation and the second approach utilizes the Alternating Direction Method of Multipliers (ADMM).
\subsection{Optimization on Stiefel Manifold} \label{OSM}
In this section, we propose optimizing  $U$  on the Stiefel manifold using the Cayley transformation as introduced in~\cite{wen2013feasible}. The Cayley transformation is a powerful tool for generating smooth, feasible curves on the Stiefel manifold by constructing orthogonal updates to  $U$. Specifically, it transforms  $U$  by left-multiplying it with $(I_n + \frac{\tau}{2} W)^{-1} (I - \frac{\tau}{2} W)$ as:  
\begin{equation}\label{Utau}
U(\tau) = (I_n + \frac{\tau}{2} W)^{-1} (I_n - \frac{\tau}{2} W) U_0,
\end{equation}
where $W$ is a skew-symmetric matrix. Since the matrix multiplication in $(I_n + \frac{\tau}{2} W)^{-1} (I - \frac{\tau}{2} W)$ is commutative for $(I_n + \frac{\tau}{2} W)^{-1}$ and $(I - \frac{\tau}{2} W)$, we can verify the Cayley transformation
 $(I_n + \frac{\tau}{2} W)^{-1} (I - \frac{\tau}{2} W)$ is an orthogonal matrix. Therefore, it is natural that $U(\tau)^T U(\tau)=I_k$ and $U(\tau)$ is a uni-variable curve on the Stiefel manifold. 

Next, we compute the derivative with respect to $\tau$ in $U(\tau)$, which is a crucial step for deriving the directional derivative required in the curvilinear search for an optimal $\tau$. By differentiating \eqref{Utau} with respect to $\tau$, we obtain:
\begin{equation}\label{Utaup}
\begin{aligned}
U'(\tau) = & - \frac{1}{2}(I_n + \frac{\tau}{2} W)^{-1} W (I_n + \frac{\tau}{2} W)^{-1} (I_n - \frac{\tau}{2} W) U_0-\frac{1}{2} (I_n + \frac{\tau}{2} W)^{-1} W U_0\\
&=-\frac{1}{2}(I_n+\frac{\tau}{2}W)^{-1} W (U_0+U(\tau)).
\end{aligned}
\end{equation}
By setting $W$ with a special skew-symmetric matrix by $W_{U_0} = \nabla_U F(U)|_{U_0} U_0^T - U_0 \nabla_U^T F(U)|_{U=U_0}$, 
we have:
\begin{equation}\label{derivU}
U'(\tau)|_{\tau=0} = -W_{U_0} U_0= -\nabla_U F(U)|_{U=U_0}+U_0\nabla_U^TF(U)|_{U=U_0} U_0.
\end{equation}
Therefore, the derivative:
\[
\frac{d F(U(\tau))}{d\tau} |_{\tau=0} = \langle U'(\tau)|_{\tau=0}, \nabla_U F(U)|_{U=U_0}\rangle = -\frac{1}{2} \|\nabla F(U)_{U=U_0} U_0^T - U_0\nabla_{U}^T F(U)|_{U=U_0}\|_{\rm F}^2 \leq 0.
\]
This implies that $U'(\tau)|_{\tau=0}$ forms a descent direction. The curvilinear search method choose a feasible $\tau'$ from two considerations: 
\begin{itemize}
\item The new point $\tau'$ should make $F(U(\tau))$ decrease with a significant amount from $F(U_0)$: $\rho_1\tau \frac{d F(U(\tau))}{d\tau}|_{\tau=0}$ ($0<\rho_1 < 1$), i.e.,
 $F(U(\tau')) \leq F(U(0))+\rho_1\tau' \frac{d F(U(\tau))}{d\tau}|_{\tau=0}$.
 \item  The decreasing speed for  $F(U(\tau))$ should also be larger than $\rho_2  \frac{d F(U(\tau))}{d\tau}|_{\tau=0} $\ $(0<\rho_2<1)$, i.e., 
 $\frac{d F(U(\tau))}{d\tau}|_{\tau=\tau'} < 0$ and   $\left |\frac{d F(U(\tau))}{d\tau}|_{\tau=\tau'}\right| \geq \rho_2 \left| \frac{d F(U(\tau))}{d\tau}|_{\tau=0}\right| $ .
\end{itemize}

Due to the continuity of $\frac{d F(U(\tau))}{d\tau}$, the two conditions outlined above are naturally satisfied as $\tau'\rightarrow 0$. Consequently, we can select an appropriate value of $\tau$ such that both conditions hold true by iteratively setting $\tau:=\frac{\tau}{2}$
starting from an initial value $\tau = \tau_0$. 

Next, we set $U_0$ to the new $U(\tau')$ and repeat the process as outlined in Algorithm \ref{alg1}. This iterative procedure generates a sequence $(\{U_1, U_2, \dots, U_\infty\})$ that converges towards the optimal solution of the original problem. Each iteration refines the solution, guiding the sequence closer to the desired optimality.

Since $(\{U_1, U_2, \dots, U_\infty\})$ is bounded, the accumulating point theorem guarantees the existence of a subsequence $\{U_{n_k}\}$ that converges to an accumulation point $U^*$. Furthermore,
the sequence $\{W_{U_{n_k}}, U_{n_k}\}$ converges to the point $\{W_{U^*}, U^*\}$, satisfying the condition $W_{U^*}U^*=\nabla_c F(U^*) =\bf 0 $. This implies that the first-order condition in \eqref{nonlinear_eign} is satisfied at the accumulating point $U^*$.

Next, we will examine several acceleration strategies to enhance the optimization algorithm, including a perturbed curvilinear search method designed to facilitate faster convergence. These approaches aim to optimize the efficiency of the algorithm, enabling it to reach solutions more rapidly and effectively.

\begin{algorithm}[t]\footnotesize \label{alg1}
\caption{Curvilinear Search for $U$:}
\label{alg:2}
\KwData{The affinity matrix: $A$ and the regularization parameter: $\lambda$, $0<\rho_1 \leq \rho_2< 1$, $\tau_0$ and tolerant parameter $\epsilon$.}
\KwResult{$U$}
Initialize $U_0$ via the eigenvectors corresponding the largest $K$ eigenvalues of $A$;\\
\While{$\|U_{k}U_{k}^T-U_{k-1}U_{k-1}^T\|_{\rm F} \geq \epsilon$}{
Set $U(0) = U_k, \tau = \tau_0, W = \nabla_U F(U)|_{U_0} U_0^T - U_0 \nabla_U^T F(U)|_{U=U_0}$;\\
$U(\tau) = (I_n + \frac{\tau}{2} W)^{-1} (I_n - \frac{\tau}{2} W) U$;\\
\While{$F(U(\tau))\geq F(U(0))+\rho_1\tau \langle \nabla_U F(U)|_{U=U(0)},U'(0) \rangle$ {\bf or} $\langle \nabla_U F(U)|_{U=U(\tau)}, U'(\tau) \rangle \geq \rho_2 \langle \nabla_U F(U)|_{U=U(0)}, U'(0) \rangle$}{
$\tau = \tau/2$;
}
Set $\tau_k = \tau$ and set ${U_{k+1}}:=U(\tau_k)$
}
\end{algorithm}
\subsubsection{A Faster Curvilinear Search Approach} \label{FCSA}
Note that the computations for both $U(\tau)$ and $U'(\tau)$ involve calculating the inverse of the matrix $I_n+\frac{\tau}{2}W\in {\mathbb R}^{n\times n}$, which is computationally intensive with a complexity of $O(n^3)$. In the following, we reduce this complexity from $O(n^3)$ to $O(K^3)$ by applying the Sherman-Morrison-Woodbury formula.

For $P = [\nabla_U F(U), U], \ Q = [U, -\nabla_U F(U)]\in {\mathbb R}^{n\times 2K}$, there is $W = PQ^T$. The Sherman-Morrison-Woodbury formula facilitates the computation of the inverse of an $n\times n$ matrix by reducing it to the inverse of a $2K\times 2K$ matrix, expressed as follows:

\begin{equation}\label{SMW}
\begin{aligned}
(I_n+\frac{\tau}{2}W)^{-1} = 
 (I_n+\frac{\tau}{2}PQ^T)^{-1}  
= I_n - \frac{\tau}{2} P (I_{2K}+\frac{\tau}{2}Q^T P)^{-1}Q^T.
\end{aligned}
\end{equation}
Substituting \eqref{SMW} into \eqref{Utau} and \eqref{Utaup}, we have $U(\tau)$ and $U'(\tau)$ can be rewritten as:
\begin{align}
U(\tau) = &U-\tau P(I_{2K}+\frac{\tau}{2}Q^T P)^{-1}Q^T U, \label{UQP}\\
U'(\tau) = &-\frac{1}{2}(I_{n} - \frac{\tau}{2} P (I_{2K}+\frac{\tau}{2}Q^T P)^{-1}Q^T)PQ^T(U+U(\tau)).
\end{align}
By directly taking the derivative respect to $\tau$ in $U(\tau)$ in equation \eqref{UQP}, we know that $U'(\tau)$ can also be equivalently expressed as:
\[
U'(\tau)= - P(I_{2K}+\frac{\tau}{2}Q^TP)^{-1}((I_{2K}+\frac{\tau}{2}Q^TP)^{-1})Q^TU.
\]

Since we only need to compute the inverse of the smaller matrix $I_{2K}+\frac{\tau}{2}Q^T P\in {\mathbb R}^{2K\times 2K}$ rather than the larger matrix $I_n+\frac{\tau}{2}W$, the computational cost for the inverse calculation can be reduced when $2K<n$.
\subsubsection{Perturbed Curvilinear Search}
\begin{algorithm}[t]\footnotesize
\caption{Perturbed Curvilinear Search for $U$:}
\label{alg:2}
\KwData{The affinity matrix: $A$ and the regularization parameter: $\lambda$, $\rho_1$, $\rho_2$, $\tau_0$ and $\epsilon$.}
\KwResult{$U$}
Initialize $U_0$ via the eigenvectors corresponding the largest $K$ eigenvalues of $A$\;
\While{$\|U_{k}U_{k}^T-U_{k-1}U_{k-1}^T\|_{\rm F} \geq \epsilon$}{
Set ${P} = [\nabla_U F(U_k), U_k], \ Q = [U_k, -\nabla_U F(U_k)],\ W = PQ^T$\;
Update $U_{k}(\tau) = U_k-\tau {P}(I_{2K}+\frac{\tau}{2}{Q}^T {P})^{-1}{Q}^T U_k$\;
\While{$F(U(\tau))\geq F(U(0))+\rho_1\tau \langle \nabla_U F(U)|_{U=U(0)},U'(0) \rangle$ {\bf or} $\langle \nabla_U F(U)|_{U=U(\tau)}, U'(\tau) \rangle \geq \rho_2 \langle \nabla_U F(U)|_{U=U(0)}, U'(0) \rangle$}{
$\tau = \tau/2$;
}
Set $\tau_k = \tau$ and generate random $R$ such that $R_{i,j}\sim {\cal N}(0,1),\forall i,j$;\\
Set $\widehat{P} = [R, U], \ \widehat{Q} = [U, -R]$ and $\delta_k = c\tau_k$;\\
Update $U_{k}(\delta_k) = U_k(\tau_k)-\delta_k \widehat{P}(I_{2K}+\frac{\delta_k}{2}\widehat{Q}^T \widehat{P})^{-1}\widehat{Q}^T U_k(\tau_k)$;\\
Set ${U_{k+1}}:=U_k(\delta_k)$.
}
\end{algorithm}
We also introduce the perturbed curvilinear search, which incorporates deliberate perturbations to help the algorithm escape local minima and improve convergence. Suppose we locate a feasible $\tau=\tau_k$ in the curvilinear search. Next, we perturb $U_k(\tau_k)$ by another Cayley transformation.

Specifically, we also modify  $U_{k}(\tau_k)$ using the Cayley transformation to ensure that the update remains on the Stiefel manifold. Instead of directly utilizing the gradient, we parameterize this modification by introducing a random matrix $R$, which introduces a controlled perturbation. The Cayley transformation preserves the orthogonality of $U_{k}(\tau_k)$, ensuring it stays on the Stiefel manifold, while the random disturbance aids in exploring the search space beyond local optima. The updated form for  $U_{k}(\tau_k)$ is given by:
\[
{U}_{k}(\delta) := U_k(\tau_k)-\delta \widehat{P}(I_{2K}+\frac{\delta}{2}\widehat{Q}^T \widehat{P})^{-1}\widehat{Q}^T U_k(\tau_k),
\]
where $\widehat{P} = [R, U_k(\tau_k)], \ \widehat{Q} = [U_k(\tau_k), -R]$. In comparison with \eqref{UQP}, this update perturbs $U_k(\tau)$ along a one-parameter curve by introducing a random direction, parameterized by $R_{i,j} \sim {\cal N}(0, 1)$. For $\delta$, instead of setting it via the binary search, we set $\delta_k \propto \tau_k$, meaning that $\delta_k$ is proportional to the updation length at $U_k$. Finally, we set ${U}_{k+1}(0):=U_{k}(\delta_k)$ and continue the $(k+1)$-th updating.

We provide the rationale for the choice of $\delta_k$, which enables an adaptive step size $\delta_k$ that responds to the magnitude of $\tau_k$. Specifically, when $\tau_k$ is large, we can set a relatively larger $\delta_k$, thereby increasing the potential to escape local optima or saddle points. Conversely, when the derivative is small—indicating proximity to a stable point—a smaller $\delta_k$ can promote stability in the convergence process. This dynamic adjustment of $\delta_k$ based on $\tau_k$ achieves a balance between exploration (escaping suboptimal stationary points) and exploitation (ensuring steady convergence to an optimal solution).

\begin{figure}[t]
\centering
\includegraphics[width=0.9\linewidth]{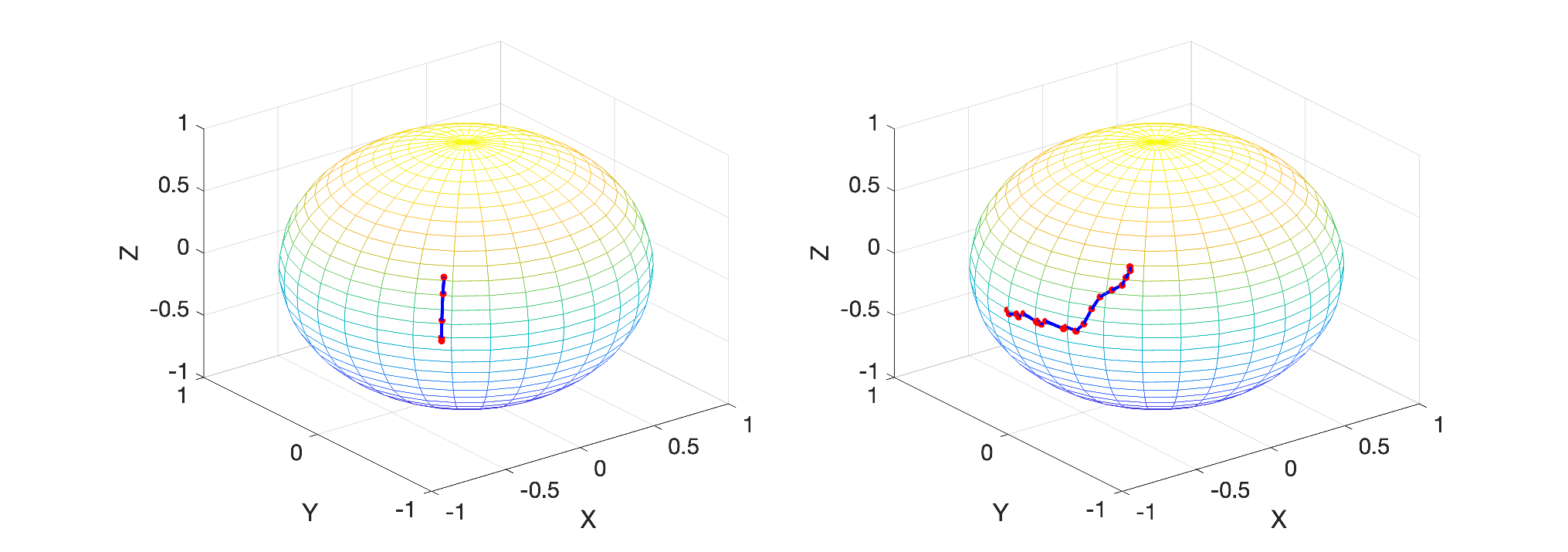}
\caption{The comparison of the convergence trajectories when applying the unperturbed (left diagram) and perturbed (right diagram) curvilnear search method for solving $\min_{\|x\|_2=1} \|x\|_1$. Both of the two methods start from $(-0.5,-0.5,0.4)$. While the unperturbed method converges to $(-\frac{\sqrt{2}}{2},-\frac{\sqrt{2}}{2},0)$ while the perturbed method converges to $(-1,0,0)$
\label{pertu_sphere}}
\end{figure}

Additionally, we provide the geometric interpretation for the perturbed curvilinear search. Similar to \eqref{derivU}, the derivative of  $U_k(\delta)$  at $\delta = 0$ is given by: 
\[
U_k'(\delta)|_{\delta=0} = -W_{U_k(\tau)} U_k(\tau)=- R+ U_k(\tau)R^T U_k(\tau),
\]
where  $R$  represents a random matrix, and $U'_k(\delta)|_{\delta=0}$ captures the random action on the tangent space under the canonical metric. 
Thus, $U_k(\delta)$, parameterized by  $\delta$ can be viewed as a perturbed point of  $U_k(\tau)$. 
If  $U_k(\tau)$ is a saddle point, such that the gradient $\nabla_U F(U)$, projected onto the tangent space  $T_{U_k(\tau)}$ equals  $\mathbf{0}$, this perturbation through  $U_k(\delta)$  offers a mechanism to help  $U_k(\tau)$  escape from the saddle point. This approach provides a controlled yet effective strategy for avoiding stagnation in non-optimal stationary points.

Finally, we present how the perturbed curvilinear search method can escape from a local optimum for the optimization problem $\min_{\|x\|_2 = 1} \|x\|_1$ discussed in Section \ref{sparse case}. Note that the constraint $\|x\|_2 = 1$ can be interpreted as a special case of an orthonormal matrix with a single column. Starting from the initial point $x_0 := (-0.5, -0.5, 0.4)$, we apply both unperturbed and perturbed curvilinear search methods.  The landscape illustrating local and global optima, shown in Figure \ref{l1l2}, provides insights into the nature of the convergence points.
Since the Cayley transformation preserves the $\ell_2$-norm of $x$, this model is well-suited to be solved using the curvilinear search approach. The convergence trajectories are shown in Figure \ref{pertu_sphere}, where we observe that the unperturbed curvilinear search approach converges to $\left(-\frac{\sqrt{2}}{2}, -\frac{\sqrt{2}}{2}, 0\right)$, whereas the perturbed approach converges to $(-1, 0, 0)$. Although both points satisfy the condition $\nabla_c F(x) = \bf 0$, the perturbed search method converges to a global optimum, while the unperturbed search method reaches only a local optimum.


\subsubsection{Derivatives for different scenarios}
Recall that in Section \ref{KKTC}, we derived the gradient $\nabla_U F(U) = -4AU+2\lambda GU$, where $G$ is a symmetric matrix that incorporates the derivatives with respect to the entries in $X:=UU^T$.
It is crucial to give the explicit forms of  $g'(\cdot)$  for three different types of penalty terms, each reflecting different regularization needs in optimization. 
For the bounded penalty, $g'_{\alpha,\beta}(z) = 2( \min\{z-\alpha,0\} - \min\{\beta-z,0\})$. For the non-negativity penalty, $g'_{0,\infty}(z)=2\min\{z,0\}$. For the sparse penalty, 
\[
g'_\delta(z) = \left\{
\begin{array}{ll}
     x/\delta,& |x|\leq \delta,  \\
     {\rm sign}(x),& \text{otherwise}.
\end{array}
\right.
\]
These different forms of the optimization problem enable the construction of the Euclidean gradient  $\nabla_U F(U) = -4AU + 2\lambda GU$  across various model settings, with  $G$  defined based on the corresponding derivatives of the objective function. 
\subsection{Optimization on Projection Matrix Manifold}
Motivated by Section \ref{OSM}, we can define a univariable curve on the rank-$K$ projection matrix manifold. Given that $X^2 = X$, we parameterize $X$ as $X(t)$ and differentiate both sides of the equation $X^2 = X$. This yields:
\begin{equation}\label{x'x}
  X'(t)X+XX'(t)=X'(t).
\end{equation}
The symmetry property of $X(t)$ implies that $X'(t)$ is also symmetric.
Multiplying $X$ on both sides of the equation $\eqref{x'x}$ gives   $XX'(t)X+XX'(t)=XX'(t)$. From this, it follows that $XX'(t)X = \bf 0$, which implies that $X'(t)$ can be expressed as:
\begin{equation}\label{X'}
X'(t) =  W(I-X)+ (I-X)W^T+(I-X)V(I-X),
\end{equation}
where $W$ is any matrix in ${\mathbb R}^{n\times n}$ matrix and $V$ is any symmetric matrix in ${\mathbb S}^{n\times n}$. Substituting $X'(t)$ from the equation \eqref{X'} into \eqref{x'x} results in the left side equaling:
\begin{equation}\label{x''}
X'(t)X+XX'(t) = (I-X)W^TX+XW(I-X).
\end{equation}
By comparing \eqref{x''} with \eqref{X'}, we obtain $W = -W^T$ and $(I-X)V(I-X)=\bf 0$, which implies that  $W$ is a skew-symmetric matrix. Consequently, we can rewrite equation \eqref{X'} as
\[
X'(t) =  XW-WX,
\]
Since $X(t)$ represents any univarible curve, it follows that $XW-WX$ spans the tangent space at $X$. Notably, the expression $(I_n + \frac{\tau}{2} W)^{-1} (I_n - \frac{\tau}{2} W)$ is an orthogonal matrix in ${\mathbb R}^{n\times n}$. Therefore, we can define a smooth curve for the rank-$K$ projection matrix as follows:
\[
X(\tau) = (I_n + \frac{\tau}{2} W)^{-1} (I_n - \frac{\tau}{2} W) X_0 (I_n + \frac{\tau}{2} W) (I_n - \frac{\tau}{2} W)^{-1},
\]
where $W\in R^{n\times n}$ is a skew-symmetric matrix. It is straightforward to verify that $X^2(\tau)=X(\tau)$ and $X(\tau)$ maintains rank-$K$. By differentiating $X(\tau)$ with respect to $\tau$, we obtain:
\[
\begin{aligned}
X'(\tau) = 
(I_n + \frac{\tau}{2} W)^{-1}[(I_n - \frac{\tau}{2} W)X_0(I_n - \frac{\tau}{2} W)^{-1}W-W(I_n + \frac{\tau}{2} W)^{-1}X_0(I_n + \frac{\tau}{2} W)](I_n - \frac{\tau}{2} W)^{-1}.
\end{aligned}
\]
Therefore, we find that $X'(\tau)|_{\tau=0}=X_0W-WX_0$. By choosing $W$ as a special skew-symmetric matrix derived from $\nabla F(X)$, defined as $W = \nabla F(X)X-X\nabla F(X)$, we obtain:
\[
\begin{aligned}
\langle X'(\tau), \nabla F(X) \rangle 
=2\|X\nabla F(X)X\|_{\rm F}^2-\|X\nabla F\|_{\rm F}^2- \|\nabla FX\|_{\rm F}^2 \leq 0.
\end{aligned}
\]
Therefore, $X'(\tau)|_{\tau=0}=X_0W-WX_0$ with $W=\nabla F(X)X-X\nabla F(X)$ provides a descent direction for $F(X)$. It is straightforward to verify that this formulation of $W$ not only serves this purpose but also acts as the coordinate for projecting $\nabla_X F(X)$ onto the tangent space of the rank-$K$ projection matrix manifold at $X$ through: 
\[
\min_{W=-W^T} \|\nabla_X F(X) - (XW-WX)\|_{\rm F}^2,
\]
which implies that $X'(\tau)|_{\tau=0}$ is the steepest descending direction on the rank-$K$ projection manifold for $F(X)$ at $X(0)$. We give our curvilinear search optimization method for $X$ in Algorithm \eqref{alg3}. 

Since both $\nabla F(X)$ and $X$ are square matrix, the inverse $(I_n+\frac{\tau}{2}W)^{-1}$ can no longer be efficiently computed using the Sherman-Morrison-Woodbury formula, as discussed in Section \ref{FCSA}. Therefore, directly applying the curvilinear search to this setting entails a higher computational cost compared to the case where optimization is performed on the orthonormal matrix $U$, which allowed for more efficient calculations.

\begin{algorithm}[t]\label{alg3} \footnotesize
\caption{Curvilinear Search for $X$:}
\label{alg:2}
\KwData{The affinity matrix: $A$ and the regularization parameter: $\lambda$, $0<\rho_1 \leq \rho_2< 1$, $\tau_0$ and tolerant parameter $\epsilon$.}
\KwResult{$X$}
Initialize $X_0$ via the eigenvectors $U_0$ corresponding the largest $K$ eigenvalues of $A$ by: $X_0 = U_0U_0^T$.\\
\While{$\|X_{k}-X_{k-1}\|_{\rm F} \geq \epsilon$}{
Set $X(0) = X_k, \tau = \tau_0$, $W = \nabla_X F(X)|_{X=X(0)}X(0)-X(0)\nabla_X F(X)|_{X=X(0)}$;\\
$X(\tau) = (I_n + \frac{\tau}{2} W)^{-1} (I_n - \frac{\tau}{2} W) X(0) (I_n + \frac{\tau}{2} W) (I_n - \frac{\tau}{2} W)^{-1}$;\\
\While{$F(X(\tau))\geq F(X(0))+\rho_1\tau \langle \nabla_X F(X)|_{X=X(0)},X'(0) \rangle$ {\bf or} $\langle \nabla_X F(X)|_{X=X(\tau)}, X'(\tau) \rangle \geq \rho_2 \langle \nabla_X F(X)|_{X=X(0)}, X'(0) \rangle$}{
$\tau = \tau/2$;
}
Set $\tau_k = \tau$ and ${X_{k+1}}:=X(\tau_k)$
}
\end{algorithm}


In addition to the curvilinear gradient search method, we explore the alternating optimization strategy for optimizing equation \eqref{Convex}. This technique decouples the projection constraint from the penalty function, enabling alternating updates of the primary variables $X$ and $Y$ through optimal solutions, rather than incremental stepwise updates. By doing so, this approach accelerates the convergence process, providing a more efficient optimization framework
\subsection{Alternating Method}
Unlike the previous approach of optimizing on the Stiefel manifold using the Cayley transformation, the Alternating Direction Method of Multipliers (ADMM) offers a distinct strategy to solve the problem in \eqref{equ}. In this approach, we separate the constraints by introducing an auxiliary variable, which allows us to decompose the problem into simpler subproblems that are easier to solve iteratively. The idea is to gradually penalize the difference between the auxiliary and the primary variables, eventually enforcing their agreement.

First, it is evident that solving the original optimization problem in \eqref{equ} is equivalent to reformulating it by introducing an auxiliary variable:
\begin{equation}\label{eqn3}
\min_{X\in\mathcal{P}_{K}, X=Y}\|A-X\|^2_{\rF}+\lambda \sum_{i,j} g(Y_{i,j}).
\end{equation}
Then, we define the augmented Lagrangian for \eqref{eqn3}:
\begin{equation}\label{objective1}
\begin{aligned}
{\cal L}_\rho (X, Y, \Lambda)  =\|A-X\|_{\rm F}^2 + \lambda\sum_{ij} g(Y_{ij}) +  \frac{\rho}{2} \|X-Y\|_{\rm F}^2 +\langle \Lambda, X-Y \rangle. 
 \end{aligned}
\end{equation}
The alternating direction method of multipliers  solves \eqref{objective1} by constructing an iterative sequence as follows:
Starting from initialization points $\{X_0,Y_0,\Lambda_0\}$, ADMM updates $\{X_k,Y_k,\Lambda_k\}$ alternatively as:
\begin{align}
X_{k+1} &=\argmin_{X\in {\cal P}_K} {\cal L}_\rho (X, Y_k, \Lambda_k),\label{X}\\
Y_{k+1} &=\argmin_{Y} {\cal L}_\rho (X_{k+1}, Y, \Lambda_k),  \label{Y}\\
\Lambda_{k+1} &= \Lambda_k + \rho(X_{k+1}-Y_{k+1}).\label{lambda}
\end{align}

These updates in \eqref{X} and \eqref{Y} have closed-form solutions and thus can be implemented efficiently. Specifically, the problem in \eqref{X} is equivalent to the following problem:
\[
X_{k+1}=\argmax_{X\in {\cal P}_K} \langle X,W_k\rangle,\quad W_k=2A+\rho Y_k-\Lambda_k.
\]
This problem of minimizing the inner product of the projection matrix can be effectively addressed by utilizing the eigenvalue decomposition of $W_k$ as follows:
\[
\widehat{U} = \max_{U^TU = I} {\rm trace}(U^T W_k U).
\]
Therefore, $X_{k+1}$ is given by the projection matrix associated with the leading $K$ eigenvectors of $W_k$, i.e., $X_{k+1} = \widehat{U}\widehat{U}^T$. 

The update with respect to $Y$ can be reformulated as an equivalent problem \eqref{Y} as: 
\begin{equation}\label{obj_un}
Y_{k+1}=\argmin_{Y}\|Y-V_{k+1}\|_{\rF}^2+\tau\sum_{ij}g(Y_{ij}),
\end{equation}
where $V_{k+1}=X_{k+1}+\Lambda_k/\rho$ and $\tau={2\lambda}/{\rho}$. This problem is separable, allowing each entry  $\{Y_{k+1}\}_{ij}$  to be addressed using the method discussed previously in relation to \eqref{univariable}.
Finally, we discuss how to solve \eqref{obj_un}.
Noticing that $\|A-X\|_{\rm F}^2 = \sum_{i,j} (A_{i,j}-X_{i,j})^2$, 
we first drop the constraint $X\in {\cal P}_K$ and rewrite \eqref{objective1} as a problem
\begin{equation}\label{appxi_entrywise}
\min_{\{X_{i,j}\}} \sum_{i,j}\{(A_{i,j}-X_{i,j})^2+\lambda g(X_{i,j})\}.
\end{equation}
The optimization problem with respect to variables $\{X_{i,j}\}$ are separable in \eqref{appxi_entrywise}, allowing the optimal minimization to be decomposed into smaller, more manageable subproblems. Consequently, it is essential to investigate the unconstrained univariate problem, formulated as follows:
\begin{equation}\label{univariable}
f(z) = (s-z)^2+\tau g(z).
\end{equation}

The convexity of $g(z)$ ensures that $\min_z f(z)$ possesses a unique optimal solution. Unraveling this solution not only yields the optimal outcome for $\min_z f(z)$ but also offers valuable insights into our regularized projection approximation problem.

\subsection{Solution to Univariate Problems}
In this section, we explore the closed-form solutions for equation \eqref{univariable} under various penalty functions $g(\cdot)$, corresponding to different settings: bounded penalty $g_{\alpha,\beta}(\cdot)$, non-negative $g_{0,\infty}(\cdot)$, and sparsity-inducing penalty $g_{\delta}(\cdot)$.
\subsubsection{Bounded Penalty \label{bounded P}}
For the bounded loss scenario, we define $g(z) := g_{\alpha,\beta}(z)$ with $\alpha<\beta$ as specified in \eqref{bb}. Subsequently, we derive the closed-form solution for $\min_{z\in \mathbb R} (s-z)^2+\tau g_{\alpha,\beta}(z)$. Substituting $g_{\alpha,\beta}$ into the regularized model leads to
\begin{equation}\label{bounded_problem}
\min_{z\in \mathbb R} \big\{ (s-z)^2+\tau \{(\min\{z-\alpha,0\})^2+(\min\{\beta-z,0\})^2\} \big\}.
\end{equation}

Depending on the magnitudes of $z$, $\alpha$, and $\beta$, the objective function takes on three distinct forms across three specific intervals. We  analyze and derive the optimal solution for each interval individually to gain a comprehensive understanding of the overall minimization problem.

{\noindent \bf Case 1}: For $z \leq \alpha < \beta$, problem \eqref{bounded_problem} simplifies to:
$\widehat{z}_1 = \arg\min_{z\leq \alpha} (z-s)^2+\tau (z-\alpha)^2.$
Therefore, the optimal solution in this range is the minimum of $\alpha$ and $\frac{s + \tau \alpha}{\tau + 1}$, which yields:
$\widehat{z}_1 = \min \{ \alpha, \frac{s+\tau \alpha}{\tau+1} \} = \frac{\tau \alpha+ \min\{\alpha, s \}}{\tau+1}$.

{\noindent \bf Case 2}: For $z \geq \beta$, the objective function in \eqref{bounded_problem} becomes:
$\widehat{z}_2 = \arg\min_{z\geq \beta} (z-s)^2 + \tau (\beta-z)^2 .$
The optimal solution here is the maximum of $\beta$ and $\frac{s+\tau \beta}{\tau+1}$, giving:
$\widehat{z}_2 = \max \{\beta,  \frac{s+\tau \beta}{\tau+1}  \} = \frac{\tau\beta +\max\{\beta,s\}}{\tau+1}$.

{\noindent \bf Case 3}:
For $\alpha \leq z \leq \beta$, in this interval, the problem \eqref{bounded_problem} reduces to:
$\widehat{z}_3 = \arg\min_{\alpha \leq z\leq \beta}  (z-s)^2 .$
Similarly to the previous cases, the optimal solution for the problem constrained by $\alpha\leq z \leq \beta$ is given by: $\widehat{z}_3 =\max\{\alpha, \min\{ \beta, s\}\}$.

\begin{table}[t]
\centering
\caption{The values of $\widehat z_1, \widehat z_2, \widehat z_3$ along with their respective function values $f(\widehat z_1), f(\widehat z_2), f(\widehat z_3)$ are provided for three distinct magnitude scenarios of $\alpha, \beta$, and $s$.}
\vspace{2mm}
\resizebox{0.6\linewidth}{!}{
    \begin{tabular}{c|ccc|ccc}  \hline \hline
    & $\widehat{z}_1$  &  $\widehat{z}_2$    & $\widehat{z}_3$   & $f(\widehat{z}_1)$ & $f(\widehat{z}_2)$ & $f(\widehat{z}_3)$ \\\hline 
   $s\leq \alpha$ & {$\frac{\tau\alpha+s}{\tau+1}$}  & $\beta$& $\alpha$ & $\frac{\tau(\alpha-s)^2}{\tau+1}$& $(\beta-s)^2$ & $(\alpha-s)^2$\\ 
   $\alpha \leq s \leq \beta$ & $\alpha$ & $\beta$ & {\bf $s$} & $(\alpha-s)^2$  &$(\beta-s)^2$ & $0$\\ 
   $\beta \leq s$     & $\alpha$&  $\frac{\tau\beta+s}{\tau+1}$ & $\beta$ &$(\alpha-s)^2$ &$\frac{\tau(\beta-s)^2}{\tau+1}$ &$(\beta-s)^2$ \\\hline \hline 
    \end{tabular}}
\label{tab:my_label_occasions}
\end{table}

To derive the global optimal solution, we consolidate the results from the three cases by comparing the optimal function values across each interval. Let $\widehat z_1,\widehat z_2$, and $\widehat z_3$ denote the optimal solutions to \eqref{bounded_problem} on three distinct intervals. Evaluating the function values at these points, $f(\widehat{z}_1), f(\widehat{z}_2), f(\widehat{z}_3)$, allows us to identify the best among them.


The optimal solutions for each interval are summarized in Table \ref{tab:my_label_occasions}. In summary, the optimal solution $\widehat{z}$ is given by:
\begin{equation}\label{solution_bound}
\widehat{z}  = \left \{
\begin{array}{cl}
  \frac{\tau \alpha+s}{\tau+1},   &  s\leq \alpha,\\
   s,  &  \alpha \leq s \leq \beta,\\
   \frac{\tau\beta+s}{\tau+1}, & \beta \leq s.
\end{array} \right.
\end{equation}

This solution can also be represented in a more compact form as:
$
\widehat{z}=\frac{1}{1+\tau} (s+\tau{P}_{\alpha,\beta}(s)),
$
where ${P}_{\alpha,\beta}(\cdot)$ is a projection operator defined by:
$
{P}_{\alpha,\beta}(s)=\min\{\max\{s,\alpha\},\beta\}
$. 
Referring to Table \ref{tab:my_label_occasions}, we observe that for $s$ outside the range $[\alpha,\beta]$, the optimal solution can be interpreted as a convex combination, $ws + (1-w)\mu$, where $w = \frac{1}{\tau+1}$ and $\mu = \arg \min_{t \in [\alpha,\beta]} |t - s|$. As $\tau$ approaches infinity, the solution converges to the projection of $s$ onto $[\alpha,\beta]$, that is, $\widehat{z} \rightarrow {P}_{\alpha,\beta}(s)$.
\subsubsection{Non-negativity Penalty}
For the non-negativity requirement, we set the penalty term as $g(z): = (\min\{z,0\})^2$. This penalty term penalizes negative values of $z$, encouraging non-negativity. Comparing this non-negativity penalty solution in \eqref{non-negative} to the bounded penalty solution in \eqref{solution_bound}, we observe that the non-negativity solution is a special case of \eqref{solution_bound}, where the lower bound is set to  $\alpha = 0$  and the upper bound to  $\beta = +\infty$. It is straight forward to derive the optimal solution yields a form by:
\begin{equation}\label{non-negative}
\arg\min_z f(z) = \left\{
\begin{array}{lc}
   \frac{s}{1+\tau},  & s\leq 0, \\
   s,  & s >0.
\end{array}
\right.
\end{equation}

From \eqref{non-negative}, we observe that $g(z)$ applies a penalty to the non-negative values of $z$ by a factor $\frac{1}{1+\tau}$, while leaving positive values of $ s $ unaffected. This aligns with the results of Section~\ref{bounded P} when setting $\alpha = 0$ and $\beta = +\infty$.


\subsubsection{Sparsity Penalty}
For the Huber loss penalty, we set $f_\delta(z):=(z-s)^2+\tau g_\delta(z)$, where $\delta$ is a hyperparameter that controls the transition point between the quadratic and absolute value components of the Huber loss.  In the following, we examine the effect of the Huber penalty $g_\delta(z)$ on the solution of $\min_z f_\delta(z)$ by considering three distinct cases, depending on the magnitude of $z$ relative to $\delta$. These cases are outlined below:

{\noindent \bf Case 1}:   For $|z| \leq \delta$, the Huber penalty behaves quadratically and the optimization problem $\min_{|z|\leq \delta} f_\delta(z)$ simplifies into:
\[
\begin{aligned}
 \widehat z_1 
              = \arg\min_{|z|\leq \delta} (1+\frac{\tau}{2\delta})(z-\frac{2\delta s}{2\delta+\tau})^2,
\end{aligned}
\]
The optimal solution can be expressed as
$\widehat z_1 = \max\{-\delta, \min\{\frac{2\delta s} {2\delta+\tau}, \delta\}\}$, which can be viewed as a projection of $\frac{2\delta s} {2\delta+\tau}$ onto the interval $[-\delta,\delta]$.

{\noindent \bf Case 2}: For $z\geq \delta$, the Huber penalty transitions into the linear function and  the optimization problem $\min_{z\geq \delta} f_\delta(z)$ simplifies into
\[
\widehat{z}_2= \arg \min_{z\geq \delta} (z-s)^2+\tau z  =\arg \min_{z\geq \delta}(z - (s-\frac{\tau}{2}))^2+s^2.
\]
Similarly, the optimal solution yields the closed-form as
$\widehat z_2:=\max\{\delta, s-\frac{\tau}{2}\}$, which is the projection of $s-\frac{\tau}{2}$ onto $[\delta,+\infty)$

{\noindent \bf Case 3}: For $z\leq -\delta$, the Huber loss function also behaves linearly and the objective function simplifies into:
\[
\widehat z_3 = \arg \min_{z\leq -\delta} (z-s)^2-\tau z  =\arg \min_{z\leq -\delta}(z - (s+\frac{\tau}{2}))^2+ s^2. 
\]
Similarly to the above two cases, this problem yields the closed-form solution as $\widehat z_3 := \min \{-\delta, s+\frac{\tau}{2}\}$ and it can be viewed as the projection of $s+\frac{\tau}{2}$ onto $(-\infty,-\delta]$.


Since we have derived the optimal solutions for $f_\delta(z)$ across three distinct intervals, the global optimal solution can be identified by comparing the function values at $\widehat{z}_1,\widehat{z}_2,\widehat{z}_3$. Given the complex relationships among $f(\widehat{z}_1),f(\widehat{z}_2),f(\widehat{z}_3)$, explicitly determining the minimum is challenging. Additionally, we list the optimal values for $\widehat{z}_1,\widehat{z}_2,\widehat{z}_3$ corresponding to five different cases depending on $s,\tau$ and $\delta$ in Table \ref{tab:my_label_solution}. 

Upon we obtain $\widehat{z}_1, \widehat{z}_2, \widehat{z}_3$, we can compute
$f_\delta(\widehat{z}_1),f_\delta(\widehat{z}_2),f_\delta(\widehat{z}_3)$, contingent upon the magnitude relationship among $s, \tau$ and $\delta$. Therefore, for any given $s,\tau$ and $\delta$, we can get the optimal value of $f_\delta(\cdot)$ by comparing $f_\delta(\widehat{z}_1),f_\delta(\widehat{z}_2)$, $f_\delta(\widehat{z}_3)$. We then set the optimal solution to $f_\delta(z)$ as the value from $\{\widehat{z}_k,k=1,2,3\}$ that achieves the smallest value of $\{f_\delta(\widehat{z}_k),k=1,2,3\}$, i.e., $\widehat{z} = \arg\min_{k=1,2,3} f_\delta (\widehat z_k)$.

\begin{table}[t]
    \centering    
    \caption{The values of $\widehat z_1, \widehat z_2, \widehat z_3$ along with their respective function values $f(\widehat z_1), f(\widehat z_2), f(\widehat z_3)$ are provided for five distinct magnitude scenarios of $\tau, \delta$, and $s$.}
    \vspace{2mm}
    \resizebox{\linewidth}{!}{
    \begin{tabular}{c|c|c|c|c|c} \hline \hline
     &  $s+\frac{\tau}{2}\leq -\delta$   & $s-\frac{\tau}{2} \leq -\delta \leq s+\frac{\tau}{2}$ & $ -\delta \leq s-\frac{\tau}{2} \leq s+\frac{\tau}{2} \leq \delta$ & $s-\frac{\tau}{2} \leq \delta \leq s+\frac{\tau}{2}$ &  $\delta\leq s-\frac{\tau}{2}$ \\ \hline
  { $\widehat{z}_1$ }  &  $-\delta$  & $\frac{2\delta s}{2\delta+\tau}$  &$\frac{2\delta s}{2\delta+\tau}$ &$\frac{2\delta s}{2\delta+\tau}$ & $\delta$ \\ 
   $\widehat{z}_2$   & $\delta$  & $\delta$ &$\delta$ &$\delta$ & $s-\frac{\tau}{2}$ \\ 
   $\widehat{z}_3$   & $s+\frac{\tau}{2}$  & $-\delta$ & $-\delta$&$-\delta$ &$-\delta$ \\ \hline
    $f(\widehat{z}_1)$   & $(\delta+s)^2+\frac{\tau\delta}{2}$  & $\frac{\tau s^2}{2\delta +\tau}$ & $\frac{\tau s^2}{2\delta +\tau}$&$\frac{\tau s^2}{2\delta +\tau}$ & $(\delta-s)^2+\frac{\tau\delta}{2}$ \\ 
    $f(\widehat{z}_2)$ & $(\delta-s)^2+\tau \delta$ &$(\delta-s)^2+\tau \delta$ &$(\delta-s)^2+\tau \delta$ & $(\delta-s)^2+\tau \delta$ & $s^2$\\ 
    $f(\widehat{z}_3)$ & $s^2$ &$(\delta+s)^2+\tau \delta$ & $(\delta+s)^2+\tau \delta$ &$(\delta+s)^2+\tau \delta$ & $(\delta+s)^2+\tau \delta$\\ \hline \hline
    \end{tabular}}
    \label{tab:my_label_solution}
\end{table}


\subsection{Convergence for ADMM}
This section provides convergence properties of the proposed ADMM algorithm. We show that any limiting point of the solution sequence is a stationary point of problem \eqref{Convex} as long as $g(\cdot)\in \cal G$.
Our proof consists of three components. 

We first establish Lemma~\ref{dual}, which states that the successive variation in the dual variable $\Lambda$ is governed by the corresponding change in $Y$, scaled proportionally by a constant factor $2\lambda \ell$. Here, $\lambda$ denotes the trade-off parameter in the objective function, while $\ell$ represents the Lipschitz constant associated with the penalty functions—specifically $g_{\alpha,\beta}(\cdot)$ for the bounded scenario or $g_\delta(\cdot)$ for the sparse case. 


\begin{lemma}\label{dual}
 $\|\Lambda_{k+1}-\Lambda_k\|_{\rF}\leq 2\lambda \ell\|Y_{k+1}-Y_{k}\|_{\rF}$.
\end{lemma}

We postpone the proof of this lemma in the Appendix. The validity of this inequality becomes apparent upon observing that $\nabla_Y {\cal L}(X_{k+1}, Y, \Lambda_k)|_{Y = Y_{k+1}} = 0$ and considering the update of the dual variable in \eqref{lambda}. Next, we show that ${\cal L}_{\rho}(X_{k},Y_{k},\Lambda_{k})$ decreases with $k$ and the difference is lower bounded by $\frac{7\ell \lambda}{6} \|Y_{k+1}-Y_k\|_{\rF}^2$ when we set $\rho=3\lambda\ell$.
\begin{lemma} \label{covergence}
Let $\rho=3\lambda\ell$. The following inequality holds:
\begin{equation}\label{L_}
{\cal L}_\rho (X_{k}, Y_k, \Lambda_k)-{\cal L}_\rho (X_{k+1}, Y_{k+1}, \Lambda_{k+1}) \notag 
\geq \frac{7\ell \lambda}{6}  \|Y_{k+1}-Y_k\|_{\rF}^2. 
\end{equation}
\end{lemma}

The proof of Lemma \ref{covergence} is provided in the appendix. The boundedness and monotonity of the sequence $\{{\cal L}_\rho(X_k,Y_k,\Lambda_k),k=1,2,\cdots\}$ imply its convergence, along with the condition $\lim_k \|Y_{k+1}-Y_k\|_{\rm F}^2\rightarrow 0$. The inequality in \eqref{L_} can be verified by applying the convexity of $g(\cdot)$ and Lemma \ref{dual}.
Finally, we demonstrate that any limiting point of $\{X_k,Y_k,\Lambda_k\}$ is a stationary point of problem \eqref{Convex}.

\begin{theorem} \label{main theorem}
Let $\rho = 3\lambda\ell$ and $\{X^*,Y^*,\Lambda^*\}$ be the limiting point of $\{X_k,Y_k,\Lambda_k\}$. Suppose $W^*=2A+\rho Y^*-\Lambda^*$ has distinct $K$-th and $(K+1)$-th eigenvalues. Then, the Karush–Kuhn–Tucker (KKT) condition of problem \eqref{Convex} holds for the limiting point, i.e.,
\[\left\{
\begin{aligned}
&(2A-\lambda g'(X^*)) U^* = U^*\Lambda_d,\\
&{U^*}^T U^* = I_d.
\end{aligned}\right.
\]
where $\Lambda_d$ is the diagonal matrix and the columns of  $U^*$ are the  eigenvectors associated with the top $K$ of $W^*$.
\end{theorem}

The proof of Theorem \ref{main theorem} is postponed to the Appendix. Theorem \ref{main theorem} can be proved by applying the KKT condition~\cite{boyd2004convex} in updates with respect to $X$ and $Y$ in \eqref{X} and \eqref{Y}, the non-negativity $g(z)$ together with the conclusion in Lemma \ref{covergence}.
\section{Numerical Experiment}
First, we analyze and compare the convergence behaviors of curvilinear search, perturbed curvilinear search, and ADMM in optimizing the model. Next, we demonstrate the effectiveness of our regularized projection approximation approach in community detection tasks using synthetic data. Finally, we validate the performance of our method on real-world datasets, comparing it against several state-of-the-art techniques to highlight its advantages.
\subsection{Comparison for Three Approaches }
We conduct a numerical experiment to illustrate the differences between the unperturbed curvilinear search, the perturbed curvilinear search, and ADMM. While analyzing the global landscape of $F(X)$ remains challenging, the experiment provides valuable insights into the convergence behavior of the two methods, highlighting their performance distinctions.

To visualize the convergence behavior of the matrices sequence $\{X_k,k = 1,2,\cdots\}$, we define a function $\phi(\cdot)$ to map $\{X_k\}$ into two-dimensional points and observe the convergence trajectory in a low-dimensional space such as in ${\mathbb R}^2$. Specifically, the function $\phi(\cdot)$ can be defined as follows: first, we select an index set $I = \{i_1,i_2,\cdots,i_{n/2}\}$ from $\{1,2,...,n\}$ and set $X_I = [X_{\cdot,i_1},...,X_{\cdot,i_{n/2}}]$, which contains a total of $n/2$ columns of $X$. We use $I^c$ to represent the complement set of $I$. Subsequently, we define a function $\phi(\cdot):{\mathbb R}^{n\times \frac{n}{2}}\rightarrow {\mathbb R}^2$ to demonstrate the convergence behavior of the iterative sequence $\{X_k,k=1,...,\infty\}$ by 
\begin{equation}\label{phi}
\phi(X) := [\frac{n^2}{2}\sum_{j\in I,i}X_{i,j}, \frac{n^2}{2}\sum_{j\in I^c,i}X_{i,j}] \in {\mathbb R}^2.
\end{equation}
Obviously, the convergence of $\{\phi(X_k),k=1,2,\cdots\}$ serves as a necessary condition for the convergence of $\{X_k,k=1,2,\cdots\}$. Thus, the convergence behavior of $\{\phi(X_k),k=1,2,\cdots\}$ provides insight into the convergence properties of $\{X_k,k=1,2,\cdots\}$, offering an indirect yet informative indication of whether $X_k$ is approaching a stable solution.

We implement the unperturbed search and perturbed search with the same initial point and observe the function value and the convergence trajectory of $\phi(X_k)$ varies with $k$.
\begin{figure}[t]
\hspace{-0.05\linewidth}
\includegraphics[width=1.1\linewidth]{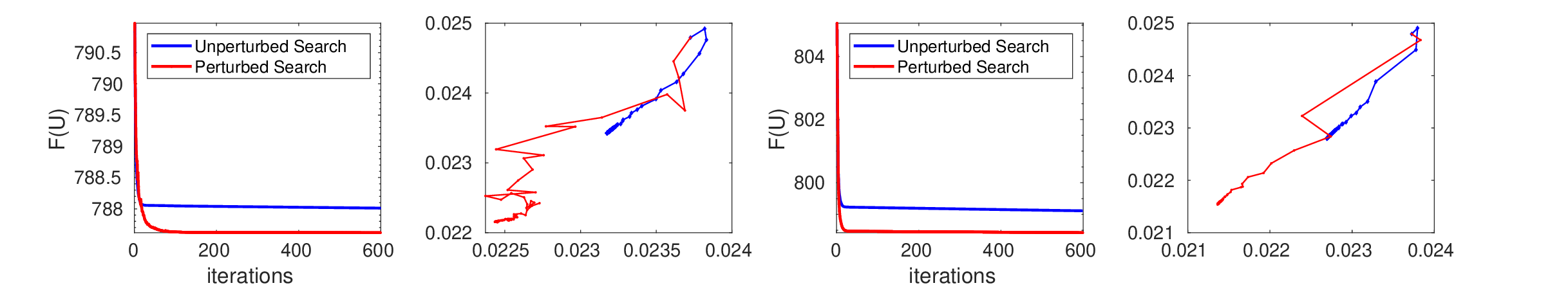}
\vspace{-1cm}
\caption{ Illustration of why the perturbed curvilinear search accelerates convergence compared to the unperturbed version: The first and third diagrams display the changes in the function value 
$F(X_k)$ across iterations $k$, while the second and fourth diagrams illustrate the convergence trajectory of $\phi(X_k)\in {\mathbb R}^2$. The first two diagrams correspond to a perturbation parameter $\lambda=0.7$, and the last two represent $\lambda=1.0$.
\label{pertu}}
\end{figure}
In Figure~\ref{pertu}, we compare the traditional curvilinear search with the perturbed curvilinear search under different settings of  $\lambda = \{0.7, 1.0\}$. 
From Figure~\ref{pertu}, it is evident that the perturbed curvilinear search can locate solutions with smaller objective function values, indicating convergence to a point of higher quality. Moreover, the trajectory of $ \phi(X_k), k = 1, 2, \dots $ shows that the perturbed search method accelerates the optimization process compared to the traditional, non-perturbed approach.

\begin{figure}[t]
{\hspace{-0.05\linewidth}
\includegraphics[width=1.1\linewidth]{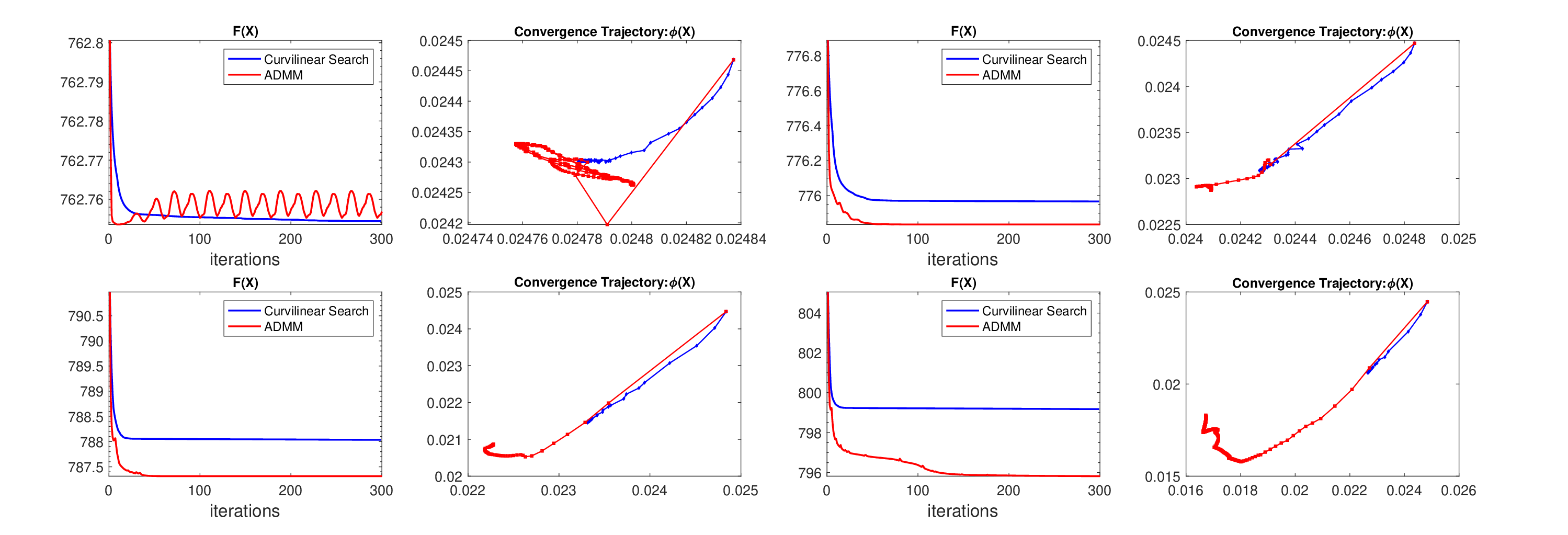}}
\vspace{-1cm}
\caption{The function value $F(X_k),k=1,2,\dots$ ($g$ select as the Huber loss penalty function) varies with the iterations $k$ and the convergence trajectory $\phi(X_k),k=1,2,\cdots$ with different set of $\lambda \in \{0.1,0.4,0.7,1.0\}$ for the curvilinear search and ADMM method. \label{admm vs curvilinear}}
\end{figure}

Additionally, we compare the performance for unperturbed curvilinear search with ADMM by starting from the same initial point. The results corresponding to different $\lambda\in\{0,1,0.4,0.7,1.0\}$ are illustrated in Figure \ref{admm vs curvilinear}. We can conclude that:
\begin{itemize}
\item While Lemma \ref{covergence} ensures the convergence of the sequence ${ {\cal L}_\rho(X_k,Y_k,\Lambda_k) },{k=1,2,\dots}$, it does not guarantee monotonic convergence for ${F(X_k)},{k=1,2,\dots}$ during the ADMM iterations. Specifically, as shown in Figure \ref{admm vs curvilinear}, we observe that $F(X_k)$ does not decrease monotonically when $\lambda$ is small. This observation highlights a potential limitation in the convergence behavior of the sequence.
\item The curvilinear search method generates a monotonically decreasing sequence of $\{F(X_k),k=1,2,\cdots\}$. However, it updates using a feasible step size $\tau$ rather than the optimal solutions with respect to $X$ and $Y$. As a result, its convergence is relatively slow compared to the ADMM, as evidenced by the convergence trajectory. This slower convergence leads to significantly higher function values, especially when $\lambda$ is large within the context of the sparse projection matrix approximation model.
\end{itemize}

Given that the curvilinear search method and ADMM follow distinct convergence trajectories, we can start with the same initial solution, such as one obtained from the eigenvalue decomposition of $X$, and apply both algorithms. By comparing their results, we can select the convergence point with the smaller function value as our final solution, thereby improving the overall outcome.


\subsection{Comparison on Clustering and Approximation}

In this section, we start with an experiment to investigate the convergence behavior across various methods. Our regularized projection matrix approximation framework encompasses several models: The bounded projection matrix approximation (BPMA) model enforces constraints to keep each element within the interval $[\alpha, \beta]$.
The positive projection matrix approximation (PPMA), a special case of the bounded projection matrix approximation (BPMA) with $\alpha = 0$ and $\beta = +\infty$.
The Sparse Projection Matrix Approximation (SPMA), which modifies PPMA by replacing its penalty term with the Huber loss function for inducing sparsity.

We begin with a synthetic example to highlight the benefits of our regularized projection approximation approach. Then we compare our proposed models—BPMA, PPMA, and SPMA—with existing community detection algorithms, including SDP-1~\cite{AMINI2018}, SDP-2~\cite{chen2016statistical}, SLSA~\cite{zhang2022graph}, and spectral clustering (SP)~\cite{von2007tutorial}, using real-world datasets. The objective functions for the methods being compared are summarized in Table~\ref{related algorithm}.

We utilize five real-world image datasets to validate the performance of the RPMA model: the Handwritten Digit dataset (DIGIT)~\cite{van1998handwritten}, the Columbia University Image Library (COIL)~\cite{Coil}, the Human Activity Recognition (HAR)~\cite{Anguita2013APD} dataset, and the Iris~\cite{misc_iris_53} and Wine~\cite{misc_wine_109} datasets from the UCI Machine Learning Repository. 
 Through this comparison, we aim to demonstrate the efficacy of our regularized projection approximation models in diverse community detection scenarios.



\begin{figure*}[t]
{\hspace{-0.05\linewidth}
\includegraphics[width=1.1\linewidth]{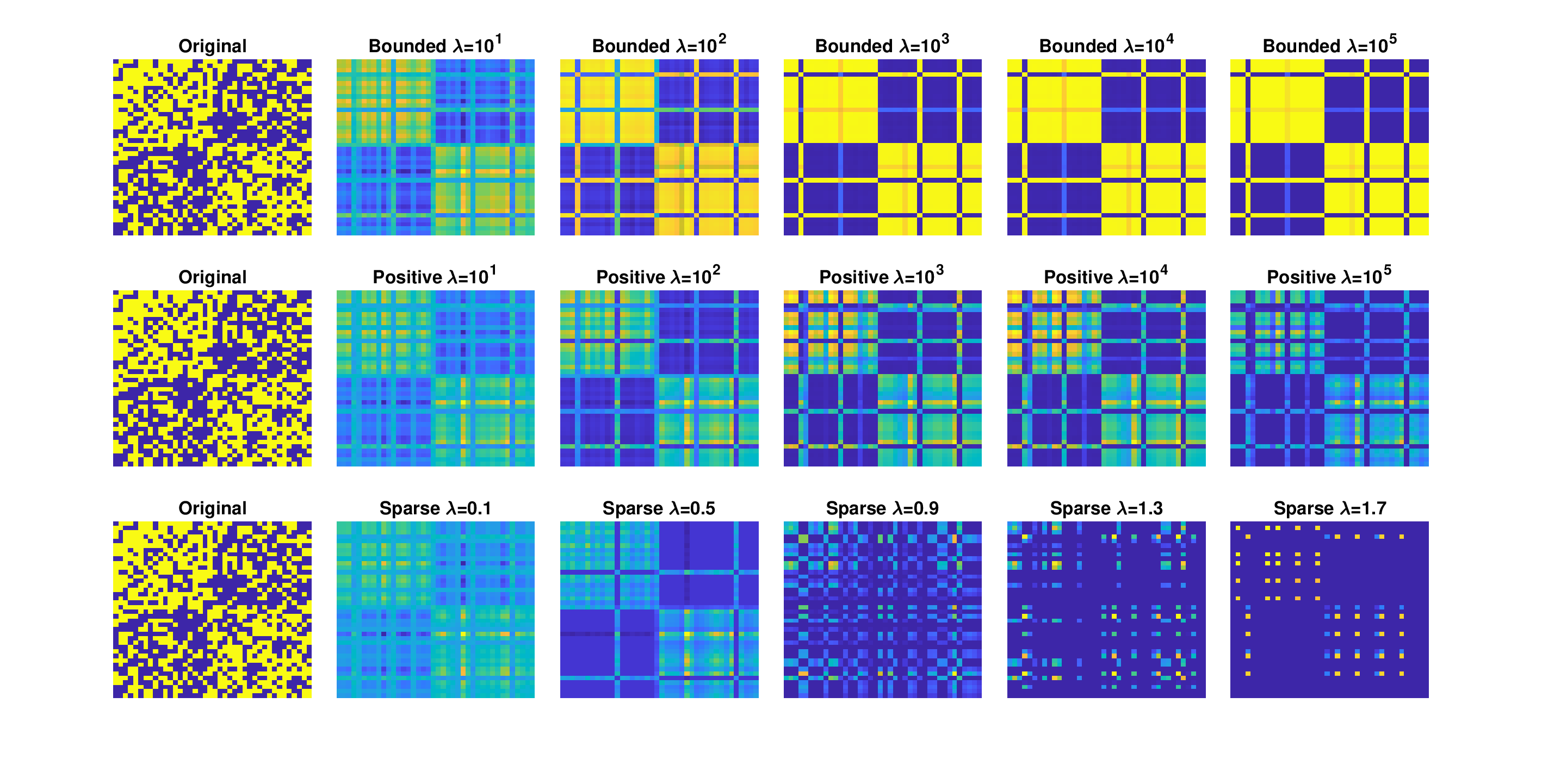}}
\vspace{-.8cm}
\caption{Illustration of solutions for the BPMA, PPMA and SPMA
with varying $\lambda$ for a synthetic dataset where $P(A_{i,j}=1)=0.65$ when $c(i)=c(j)$ and $P(A_{i,j}=1)=0.40$ when $c(i)\neq c(j)$.\label{fig:sparse_lambda}}
\end{figure*}


\begin{table}[t]
\centering
\renewcommand{\arraystretch}{2} 
\caption{The objectives and constraints for related algorithms\label{related algorithm}}
\vspace{1mm}
\resizebox{\linewidth}{!}{
\begin{tabular}{c|c|c|c|c|c}\hline \hline
&\Large SDP-1&\Large SDP-2&\Large Spectral&\Large SLSA & \Large RPMA\\ \hline
 \Large Objectives &\Large $\max_X \ \langle A, X\rangle $ & \Large $\max_X \ \langle A, X\rangle$ & \Large $\max_X \ \langle A, X\rangle$ 
&\Large  $
\min_{X,U}\|X-A\|_{\rm F}^2+\theta\|X-UU^T\|_{\rm F}^2
$ & \Large $ \min_X \|A-X\|_{\rF}^2+\lambda \sum_{ij}g(X_{ij})$
 \\ 
 \hline
\Large Constraints &\large $
\begin{aligned}
&\ X {\bf 1}_n = n/K {\bf 1}_n,  \\
&\ {\rm diag}(X) = {\bf 1}_n,  \\
&\ X\succeq 0,\ X\geq 0
\end{aligned}
$
& \large
 $
\begin{aligned}
&\ \langle X, E_n \rangle = n^2/K, \\
&\ {\rm trace}(X) = n,  \\
&\ X\succeq 0, \ 0 \leq X \leq 1
\end{aligned}
$
&\Large
$
\begin{aligned}
 &\ X\in {\cal P}_K\\ 
 \end{aligned}
$
& \Large
$
\begin{aligned}
&U^TU=I_K,\\
&\Large \|X_{\rm off}\|_0 \leq \eta
\end{aligned}
$ & \Large $X\in {\cal P}_K$
\\ \hline \hline
\end{tabular}}

\end{table}

\subsection{Experiment with Synthetic Data}

\begin{table*}[t]
\centering
\caption{Comparison of the performance of bounded, positive and sparse regularized projection matrix approximation under different setting of parameters. }
\vspace{2mm}
\renewcommand{\arraystretch}{1.5} 
\resizebox{\linewidth}{!}{
\begin{tabular}{c|c|ccccc|ccccc|ccccc} \hline \hline
&  & \multicolumn{5}{c|}{\Large Frobenius norm: $\|P_A-P^*\|_{\rm F}$}& \multicolumn{5}{c|}{\Large ACC} & \multicolumn{5}{c}{\Large NMI} \\ \hline
&\Large$\delta$  & \Large$\lambda=10$ & \Large$\lambda=10^2$ & \Large$\lambda=10^3$ & \Large$\lambda=10^4$ &\Large$\lambda=10^5$ &\Large $\lambda=10$ & \Large$\lambda=10^2$ & \Large$\lambda=10^3$ & \Large$\lambda=10^4$ &\Large$\lambda=10^5$ & \Large$\lambda=10$ &\Large $\lambda=10^2$ &\Large $\lambda=10^3$ &\Large $\lambda=10^4$ &\Large$\lambda=10^5$   \\ \hline
\Large Bounded& -- & \Large0.6938 &\Large 0.8340 & \Large0.8531 &\Large0.8531 & \Large0.8523 &\Large 0.9100 & \Large0.8750 &\Large 0.9000 &\Large 0.9000 & \Large0.9000 & \Large0.5670 & \Large0.4592 &\Large 0.5310 &\Large0.5310 &\Large0.5310 \\
\hline 
\Large Positive&-- & \Large 0.7852 & \Large0.7621 &\Large 1.0080 &\Large 1.0080 &\Large1.0092 &\Large 0.8750 &\Large 0.9000 &\Large0.8500 &\Large 0.8500 & \Large0.8250 & \Large0.4592 &\Large 0.5310 &\Large 0.3988 &\Large 0.3988 & \Large0.3464 \\\hline\hline

\multirow{6}{*}{\Large Sparse}&\Large$\delta$  &\Large $\lambda=0.1$ & \Large$\lambda=0.2$ & \Large$\lambda=0.3$ &\Large $\lambda=0.4$ &\Large$\lambda=0.5$ &\Large$\lambda=0.1$ & \Large$\lambda=0.2$ & \Large$\lambda=0.3$ & \Large$\lambda=0.4$ &\Large$\lambda=0.5$ & \Large$\lambda=0.1$ &\Large$\lambda=0.2$ &\Large $\lambda=0.3$ & \Large$\lambda=0.4$ &\Large$\lambda=0.5$ \\ \cline{2-17}

&\Large$10^{-3}$&\Large 0.8713 & \Large0.9156 &\Large 0.9084 &\Large 0.9458 &\Large 0.7307 &\Large 0.8250 &\Large 0.8313 & \Large0.8750 &\Large 0.8750 &\Large 0.9500 &\Large 0.3464 & \Large0.3470 &\Large0.4592 &\Large 0.4592 &\Large 0.7136 \\
&\Large$10^{-4}$&\Large 0.8791 &\Large0.9386 &\Large0.9026 &\Large0.9421 & \Large0.7278 &\Large 0.8250 &\Large 0.8000 & \Large0.8750 &\Large 0.8750 &\Large0.9500 & \Large0.3464 & \Large0.2832 &\Large0.4592 & \Large0.4592 &\Large 0.7136 \\
&\Large$10^{-5}$&\Large 0.8515 &\Large 0.8965 &\Large0.9025 & \Large0.9419 & \Large0.7276 &\Large 0.8638 &\Large0.8500 & \Large0.8750 &\Large0.8750 &\Large 0.9500 & \Large0.4320 & \Large0.3902 & \Large0.4592 & \Large0.4592 &\Large 0.7136 \\
&\Large$10^{-6}$&\Large0.8500 &\Large 0.8934 &\Large 0.9025 &\Large 0.9419 &\Large 0.7275 &\Large 0.8650 & \Large0.8500 &\Large 0.8750 &\Large 0.8750 & \Large0.9500 & \Large0.4350 & \Large0.3902 &\Large 0.4592 &\Large 0.4592 &\Large 0.7136 \\
&\Large$10^{-7}$& \Large 0.8498 &\Large0.8931 &\Large 0.9025 & \Large0.9419 &\Large 0.7275 &\Large 0.8600 & \Large0.8500 &\Large 0.8750 & \Large0.8750 & \Large0.9500 &\Large 0.4229 &\Large0.3902 & \Large0.4592 &\Large 0.4592 &\Large 0.7136 \\\hline \hline
    \end{tabular}}
    \label{tab:my_label_synthetic}
\end{table*}

We generate a random symmetric community connection matrix $ A $ with probabilities $ P(A_{i,j} = 1) = p_1 $ if $ c(i) = c(j) $, and $ P(A_{i,j} = 0) = p_2 $ if $ c(i) \neq c(j) $. To denoise $ A $, we apply the projection matrix approach with bounded, positive, and sparse regularizations, resulting in a regularized projection matrix $ P_A $. By rearranging elements using clustering information, we observe that $ \mathbb{E}(A) $ is a diagonal block matrix up to a permutation matrix $ Q $. We then derive the ideal rank-$ K $ projection matrix $ P_* $ by solving 
$
\min_{X \in {\cal P}_K} \| \mathbb{E}(A) - X \|_{\rm F}^2
$
and measure the distance between the regularized solution $ P_A $ and $ P_* $. Next, we apply $ k $-means clustering on the rows of the orthonormal matrix $ U \in \mathbb{R}^{n \times K} $, where $ U $ is derived from the factorization $ P_A = UU^T $ with $ U^T U = I_K $. Clustering performance is assessed using two standard metrics: accuracy (ACC) and normalized mutual information (NMI)~\cite{maulik2002performance}.


We set $p_1 = 0.65, p_2 = 0.40$ and the number of samples $n=40$ and the number of communities $K=2$. In Figure~\ref{fig:sparse_lambda}, we illustrate the solutions for BPMA, PPMA and SPMA with varying $\lambda$ ranges in $\{0.1,0.5,\cdots,1.7\}$.
For the bounded regularization technique, we set parameters $\alpha = 0$ and $\beta = 1/20$. Table \ref{tab:my_label_synthetic} displays the spectral clustering results: ACC = 0.815, NMI = 0.328, and Frobenius norm $\|P_A-P_*\|_{\rm F} = 0.898$.

We can draw the following conclusions.
First, regularization significantly enhances clustering performance, notably increasing ACC from 0.815 to 0.950 and NMI from 0.328 to 0.714 in the case of SPMA.
Second, choosing a sufficiently large value for $\lambda$ is effective for BPMA and PPMA, simplifying parameter tuning. However, for SPMA, fine-tuning $\lambda$ is essential to achieve optimal performance.
Third, SPMA improves the clustering accuracy by effectively capturing additional structural information encoded within the projection matrix, which corresponds to the true assignment matrix.


\subsection{Comparison of Clustering on Real-world Dataset}
We utilize three real-world image datasets to validate the performance of the regularized projection approximation model: the Handwritten Digit dataset (DIGIT), the Columbia University Image Library (COIL), and the Human Activity Recognition (HAR) dataset from the UCI Machine Learning Repository. 
Firstly, we evaluate our algorithm against competitors using subsets of each dataset, denoted as DIGIT-5, COIL-10, and HAR-3, which contain five, ten, and three classes, respectively. Subsequently, we conduct experiments on the full datasets, referred to as DIGIT-10, COIL-20, HAR-6, Iris, and Wine, comprising ten, twenty, six, three, and three classes, respectively.

We demonstrate how the regularized projection matrix approximation models including BPMA, PPMA and SPAM can be used to enhance spectral clustering. We compare the results corresponding to BPMA, PPMA and SPAM with SDP-1, SDP-2, Spectral clustring and SLSA on the Coil, handwritten digit, and Human Activity Recognition datasets. The similarity matrix $A$ is constructed using the Gaussian kernel such that $A_{i,j} = \exp(-\|x_i-x_j\|_2^2/\sigma^2)$ where the bandwidth parameter is set as $\sigma^2= \frac{2}{n(n-1)}\sum_{i<j}\|x_i-x_j\|_2^2$. We set the parameters $\delta\in \{10^{-3},10^{-4},10^{-5},10^{-6}\}$ and $\lambda \in \{0.1\times k,k=1,2...,8\}$ in the SPMA. For SPMA, PPMA, and BPMA, we use the result from spectral clustering as the initialization.

\begin{figure*}[t]
\hspace{-0.05\linewidth}
\includegraphics[width=1.1\linewidth]{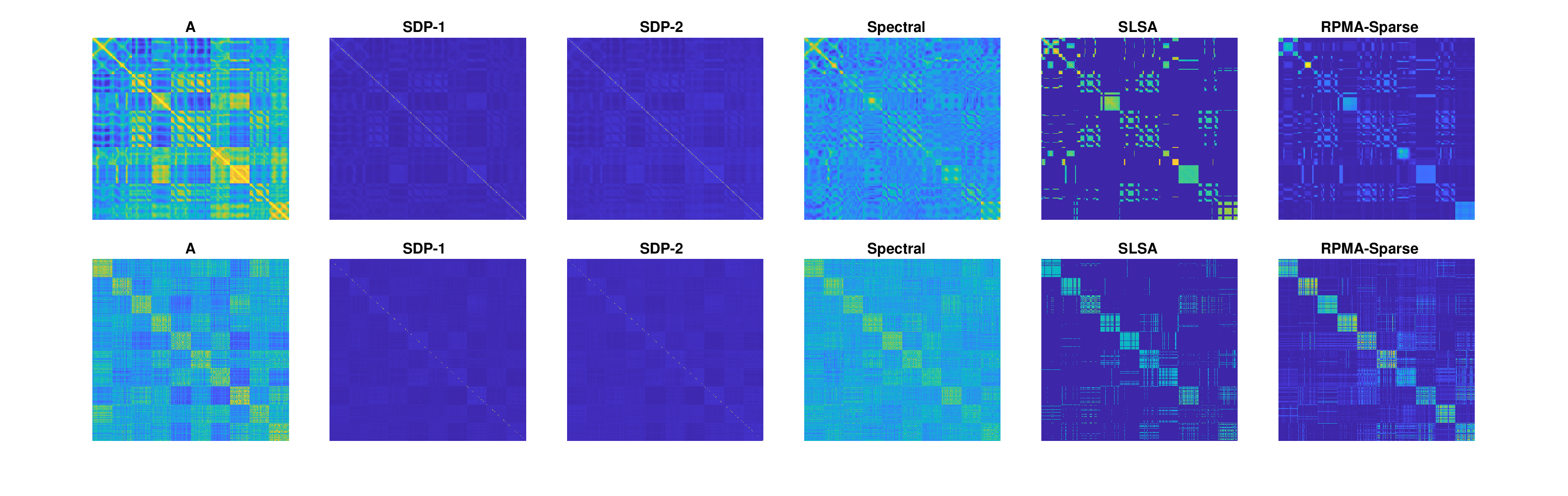}
\caption{Illustration of solutions corresponding to SDP-1, SDP-2, Spectral Clustering, SLSA, and SPMA for the COIL-10 (top row) and DIGIT-10 (bottom row) datasets, respectively. \label{solution_illu}}
\end{figure*}

Table~\ref{real_exp} summarizes the optimal performance metrics, while Figure~\ref{solution_illu} provides a visualization of the solutions produced by each algorithm. Notably, due to the non-negativity constraints in SDP-1 and SDP-2, the block diagonal elements exhibit no clear differentiation. Comparing the solutions of SLSA and SPMA in Figure~\ref{solution_illu}, we observe that SPMA accurately recovers the signal within each diagonal block, whereas SLSA suffers from significant signal degradation, especially in the 9th diagonal block. Furthermore, as shown in Table~\ref{real_exp}, SPMA achieves the highest ACC and NMI scores, demonstrating superior performance over the competing algorithms.


\begin{table}
\centering
\caption{Performance Comparison of the Related Community Detection Algorithms}
\label{real_exp}
\vspace{2mm}
\resizebox{1\linewidth}{!}{
\begin{tabular}{c|cccc|ccc|cccc|ccc}
\hline \hline
Criterion & \multicolumn{7}{c|}{ACC} & \multicolumn{7}{c}{NMI} \\
\hline
\multirow{2}{*}{Methods} & \multirow{2}{*}{SDP-1} & \multirow{2}{*}{SDP-2} & \multirow{2}{*}{SP} & \multirow{2}{*}{SLSA} & \multicolumn{3}{c|}{RPMA} & \multirow{2}{*}{SDP-1} & \multirow{2}{*}{SDP-2} & \multirow{2}{*}{SP} & \multirow{2}{*}{SLSA} & \multicolumn{3}{c}{RPMA} \\
\cline{6-8}
\cline{13-15}
& & & & & Sparse & Bound & Positive & & & & & Sparse & Bound & Positive \\
\hline
Iris & 0.887	& 0.876	  & 0.873  	& 0.879  &\bf 0.900 	 & 0.893  	& 0.880   & 0.742  	& 0.724  	& 0.721  	& 0.716  &\bf 0.758  	& 0.733  	& 0.735    \\
Wine & 0.696  & 0.689  & 0.689  & 0.702    & 0.701  & \bf0.706  & 0.631  & 0.422  & 0.426  & 0.426  & 0.423    &\bf 0.427  & 0.404  & 0.368  \\
HAR3& 0.640  & 0.638  & 0.643  & 0.642    & \bf0.701  & 0.664  & 0.651  & 0.606  & 0.605  & 0.612  & 0.554   &\bf 0.640  & 0.591  & 0.586   \\
HAR6 & 0.617  & 0.622  & 0.636  & 0.593    &\bf 0.685  & 0.621  & 0.633  & 0.609  & 0.611  & 0.614  & 0.515    &\bf 0.631  & 0.572  & 0.603   \\
DIGIT5 & 0.922 & 0.919 & 0.929 & 0.928 & \bf 0.952 & 0.938 & 0.924 & 0.818 & 0.816 & 0.819 & 0.818 & \bf 0.877 & 0.832 & 0.813 \\
DIGIT10 & 0.677 & 0.687 & 0.684 & 0.680 & \bf 0.817 & 0.634 & 0.646 & 0.635 & 0.644 & 0.643 & 0.620 & \bf 0.751 & 0.556 & 0.588\\
COIL10 & 0.541 & 0.550 & 0.550 & 0.546 & \bf 0.602 & 0.560 & 0.556 & 0.596 & 0.605 & 0.605 & 0.570 & \bf 0.654 & 0.576 & 0.604 \\
COIL20 & 0.644 & 0.625 & 0.635 & 0.569 & \bf 0.709 & 0.565 & 0.615 & 0.768 & 0.762 & 0.764 & 0.697 & \bf 0.806 & 0.677 & 0.738 \\
\hline \hline
\end{tabular}
}
\end{table}

\section{Conclusion}
We explore the regularized projection approximation framework, developing both the curvilinear search algorithm and the ADMM algorithm for its solution. This framework accommodates various models—such as bounded, nonnegative, and sparse projection matrix approximations—by adapting to different forms of the penalty function. We present the first-order optimality conditions along with a geometric interpretation. Additionally, we prove the convergence property for ADMM and demonstrate that both algorithms converge to the first-order optimal conditions. Numerical experiments highlight the effectiveness of the regularization approach on both synthetic and real-world datasets. 

Our work also has some limitations. First, while setting $\lambda$ as a sufficiently large value enforces the penalty for bounded and nonnegative scenarios, the sparsity-regularized model requires more careful tuning of $\lambda$, as an overly sparse solution can hinder the algorithm’s performance in estimating an accurate affinity matrix. Second, due to the optimization being performed over matrices rather than vector solutions, deriving the second-order optimality condition poses significant challenges, and we did not verify the convexity property at the stationary point. Additionally, we did not investigate the global landscape for optimal solutions. These aspects are left as directions for future research.


Several aspects merit future research. First, we have not yet explored the global solution landscape or how to develop algorithms to identify a global solution. Second, we believe that certain structural properties can enhance the approximation of an affinity matrix, leading to improved clustering precision; additional structures should also be investigated. Third, it is important to explore methods for accelerating the curvilinear search process. Lastly, as our framework can be generalized to models constrained on the Stiefel manifold, the application of Stiefel manifold techniques in machine learning presents another promising area for future study.


\section*{Acknowledgment}
This research was supported by the National Natural Science Foundation of China under Grants 12301478 and 11801592, as well as by the research start-up funding (Grant 312200502524) from Beijing Normal University.
\bibliographystyle{unsrt}
\bibliography{Cmm}

\end{document}